\documentclass[10pt,twocolumn,letterpaper]{article}

\usepackage{iccv}
\usepackage{times}
\usepackage{epsfig}
\usepackage{graphicx}
\usepackage{amsmath}
\usepackage{amssymb}

\usepackage{bm}
\usepackage[export]{adjustbox}
\usepackage{amsmath}
\usepackage{dsfont}

\usepackage{algorithm} 
\usepackage{algpseudocode}
\usepackage{subcaption}
\usepackage{bbm}
\usepackage[table]{xcolor}
\usepackage{colortbl}
\usepackage{soul}
\DeclareMathOperator*{\argmin}{arg\,min}
\DeclareMathOperator*{\argmax}{arg\,max}
\usepackage{booktabs}
\usepackage{dblfloatfix}    % To enable figures at the bottom of page

\usepackage[pagebackref=true,breaklinks=true,letterpaper=true,colorlinks,bookmarks=false]{hyperref}

\usepackage[capitalize]{cleveref}
\crefname{section}{Sec.}{Secs.}
\Crefname{section}{Section}{Sections}
\Crefname{table}{Table}{Tables}
\crefname{table}{Tab.}{Tabs.}

\iccvfinalcopy % *** Uncomment this line for the final submission

\def\iccvPaperID{7566} % *** Enter the ICCV Paper ID here

\ificcvfinal\fi

\begin{document}

%%%%%%%%% TITLE
%\title{Adaptive Similarity Bootstrapping for Self-Distillation}
\title{Adaptive Similarity Bootstrapping for Self-Distillation based \\Representation Learning}

\author{
Tim Lebailly$^{1*}$ \hspace{1cm} Thomas Stegm\"uller$^{2*}$ \hspace{1cm} Behzad Bozorgtabar$^{2,3}$ \\
Jean-Philippe Thiran$^{2,3}$ \hspace{1cm} Tinne Tuytelaars$^{1}$ \\ 
$^{1}$KU Leuven \hspace{1cm} $^{2}$EPFL \hspace{1cm} $^{3}$CHUV \\ 
{\small $^{1}$\texttt{\{firstname\}.\{lastname\}@esat.kuleuven.be} \hspace{2cm} $^{2}$\texttt{\{firstname\}.\{lastname\}@epfl.ch}}
}

\maketitle
\ificcvfinal\thispagestyle{empty}\fi

%%%%%%%%% ABSTRACT
\begin{abstract}
Most self-supervised methods for representation learning leverage a cross-view consistency objective \ie they maximize the representation similarity of a given image's augmented views. Recent work NNCLR goes beyond the cross-view paradigm and uses positive pairs from different images obtained via nearest neighbor bootstrapping in a contrastive setting. We empirically show that as opposed to the contrastive learning setting which relies on negative samples, incorporating nearest neighbor bootstrapping in a self-distillation scheme can lead to a performance drop or even collapse. We scrutinize the reason for this unexpected behavior and provide a solution. We propose to adaptively bootstrap neighbors based on the estimated quality of the latent space. We report consistent improvements compared to the naive bootstrapping approach and the original baselines. Our approach leads to performance improvements for various self-distillation method/backbone combinations and standard downstream tasks. Our code is publicly available at \texttt{\color{magenta}{https://github.com/tileb1/AdaSim}}.
% We empirically show that such scheme stabilizes the training and leads to performance improvements both on standard downstream $k$-NN/linear evaluations and few-shot transfer. 

%\ie they maximize the similarity between representations of two augmentations of the same image.

% This gives features which are invariant to standard data augmentation. 

% In the presence of an oracle outputting valid positive pairs (\ie pairs of images with similar semantic content), the self-distillation learning scheme could be improved. Using positive pairs from the oracle would result in features which are not only invariant to data augmentations but invariant to everything but the semantics of the image. We propose to proxy such oracle using the structure of the latent space in a self-distillation setting. Empirically, we observe that straightforward nearest neighbor bootstrapping can be hurtful and lead to collapse when used in conjunction with a self-distillation objective. 

% As opposed to previous work, we do not propose a scheme which deterministically bootstraps a nearest neighbor but instead adaptively bootstrap a neighbor sampled from a set of ranked neighbors based on their estimated semantic similarity. 

% 1) bootstrapping / constrastive works
% 2) self distillation is good (but doesn't work with nn bootstrapping)
% 3) we propose adaptive
% between the augmented views of a given image.
% 4) improvements
\end{abstract}
\newcommand\blfootnote[1]{%
  \begingroup
  \renewcommand\thefootnote{}\footnote{#1}%
  \addtocounter{footnote}{-1}%
  \endgroup
}
\blfootnote{* denotes equal contribution.}

\newcommand{\TODO}[1]{{\color{red} {\bf TODO:} #1}}

\newcommand{\aug}{\tilde{\boldsymbol{x}}}
\newcommand{\im}{\boldsymbol{x}}
\newcommand{\weight}{\boldsymbol{\theta}}

\newcommand{\rep}{\boldsymbol{z}}
\newcommand{\avgp}{\bar{\boldsymbol{z}}}

\newcommand{\glob}{\boldsymbol{z}}
\newcommand{\dense}{\boldsymbol{z}}
\newcommand{\loc}{\boldsymbol{z}^k}

\newcommand{\p}{\boldsymbol{p}}

\newcommand{\pos}{\boldsymbol{e}}

\newcommand\norm[1]{\left\lVert#1\right\rVert}

\newcommand{\augvec}{\boldsymbol{w}}

\newcommand{\augvecspa}{\boldsymbol{w}_{geo}}
\newcommand{\match}{\leftrightarrow}

\newcommand{\xmark}{\ding{55}}

\definecolor{darkgreen}{rgb}{0, 0.5, 0}

\newcommand{\Z}{\mathbf{Z}}

\newcommand{\collapse}{\textcolor{red}{-}}

% Colors
%#c6effc
%\definecolor{light_cyan}{rgb}{0.85,0.85,0.85}
\definecolor{light_cyan}{HTML}{c6effc}
\sethlcolor{light_cyan}

%%%%%%%%%%%%%%%%% REBUTTAL
\newcommand{\Raa}[1]{\textcolor{red}{#1}}
\newcommand{\Ra}{\Raa{R1}}
\newcommand{\Rbb}[1]{\textcolor{green}{#1}}
\newcommand{\Rb}{\Rbb{R2}}
\newcommand{\Rcc}[1]{\textcolor{blue}{#1}}
\newcommand{\Rc}{\Rcc{R4}}

% Define annotation commands
\newcommand{\Tim}[1]{\textcolor{magenta}{#1}}
\newcommand{\Thomas}[1]{\textcolor{blue}{#1}}
\newcommand{\Behzad}[1]{\textcolor{red}{#1}}
\newcommand{\Tinne}[1]{\textcolor{green}{#1}}
%%%%%%%%% BODY TEXT

\begin{figure}[t]
    \centering
    \includegraphics[width=\columnwidth]{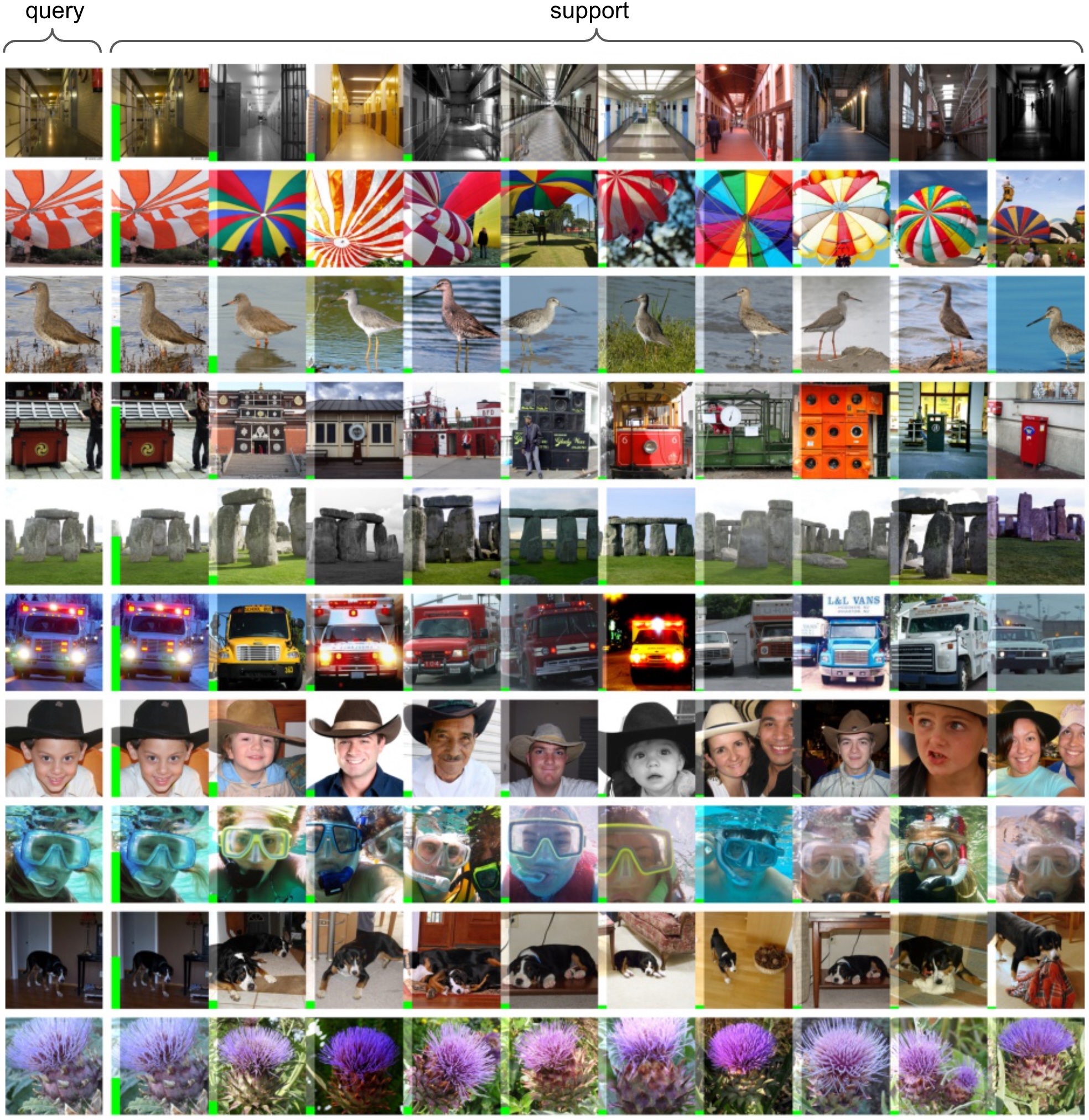}
    \caption{\textbf{The selection of positive image pairs used for cross-view consistency in self-supervised representation learning is key for good performance.} With our method, given the query (or anchor) image on the left, similar images are successfully ranked according to $p^{\text{win}}(\im_j | \im_i)$ (illustrated as a \textcolor{green}{green} bar on the bottom left of each image). Our algorithm enforces similarity between the query $\im_i$ and an image $\im_j$ sampled from $p^{\text{win}}(\im_j | \im_i)$. These results are non-cherry-picked and obtained at the final epoch (800) of the pretraining. Best viewed in color and zoomed-in.}
        % \caption{\textbf{Query image $\im_i$ (left column) and corresponding support $\mathcal{S}_{\text{win}}$ ranked by decreasing $p^{\text{win}}(\im_j | \im_i)$ (illustrated as a \textcolor{green}{green} bar on the bottom left of each image).} These results are non-cherry-picked and obtained at the final epoch (800) of the pretraining. Our algorithm enforces similarity consistencies between the query $\im_i$ and an image $\im_j$ sampled from $p^{\text{win}}(\im_j | \im_i)$. Best viewed in color and zoomed-in.}
    \label{fig:query_support}
\end{figure}

\section{Introduction}
\label{sec:introduction}
Self-supervised learning (SSL) methods have seen a lot of breakthroughs over the past few years. Most recent self-supervised methods train features invariant to data augmentation by maximizing the similarity between two augmentations of a single input image. However, this task is ill-posed as this optimization procedure admits trivial solutions (resulting in a ``collapsed'' scenario). Similarity maximization (or cross-view consistency) SSL methods can be categorized based on how they avoid trivial solutions. The most famous subset are contrastive learning methods \cite{simclrv2,pmlr-v119-chen20j,moco,mocov2,mocov3} in which the collapse is avoided by using negative pairs. On the one hand, the learning procedure is robust since collapse avoidance is explicitly modeled in the training objective, but on the other hand, it requires large batches to have a sufficient pool of negative samples. This makes them GPU memory inefficient and limits research to those who dispose of large distributed computing infrastructure.

% \begin{figure}[t]
%     \centering
%     \includegraphics[width=0.8\columnwidth]{figures/Screenshot 2022-11-02 at 16.48.47.jpg}
%     \caption{\TODO{this is only a placeholder to estimate size}}
%     \label{fig:overview}
% \end{figure}

More recently, self-distillation methods have been gaining traction \cite{simsiam,dino,byol,esvit}. These similarity maximization algorithms avoid trivial solutions by using asymmetry. This asymmetry can take the form of an additional predictor \cite{simsiam,byol} on one branch, using stop-gradients \cite{simsiam,dino,esvit,byol}, a momentum encoder \cite{dino,byol}, \etc. These methods are of particular interest as 1) they do not require large batch sizes, and 2) they currently show state-of-the-art performance \cite{dino,esvit} on standard downstream tasks.

%This oracle would yield diverse and complex pairs of images.
Orthogonal to the choice of the framework (contrastive vs self-distillation), one can wonder what is the best way to obtain positive pairs. Intuitively, similarity maximization SSL methods could be improved by using positive pairs from different images. Indeed if an oracle indicating valid positive pairs \cite{supcon,improve_supcon} was available, instead of taking two augmentations from the same image, we could simply take pairs from the oracle. The features would, therefore, not be trained to be invariant to handcrafted data augmentations but invariant to intra-class variation, which would make them more aligned with most common downstream tasks, \eg classification.

In the absence of labels, we can leverage the structure of the latent space to obtain a proxy for the oracle. Semantically related images are expected to lie in the vicinity of one another in the latent space. However, this is a chicken and egg problem, as this assumption only holds when the quality of the learned latent space is good enough. If the learned latent space is not of good quality, bootstrapping the proxy leads to unwanted gradient flows, e.g., an image of a cat is pulled closer to the image of a building.

% This idea of bootstrapping nearest neighbors is essentially orthogonal to the choice of loss-function used (contrastive or self-distillation). 

% Similarly, used in conjunction with self-distillation methods, one might expect nearest neighbor bootstrapping for positive image pair generation to be beneficial. \textbf{However, we empirically observe that straightforward bootstrapping can be hurtful and can even lead to collapse.}

% To overcome the above issue, we propose to estimate the quality of the latent space and only use neighbors as positive image pairs if the quality of the latent space is judged to be high enough. If that condition is not satisfied, we default to using standard positive pairs \ie two augmentations from the same image. 

%Recent work NNCLR \cite{dwibedi_little_2021} bootstraps nearest neighbors in a straightforward way which leads to performance improvements when used in conjunction with a contrastive objective. It then makes sense to try and combine the best solution for selecting positives pairs \cite{dwibedi_little_2021} with the best solution for defining the loss-function, arguably, self-distillation.

Nevertheless, recent work NNCLR \cite{dwibedi_little_2021} has successfully incorporated nearest neighbor (NN) bootstrapping in a contrastive setting. Considering that self-distillation typically outperforms contrastive methods, in this work, we explore how the same can be achieved without explicit use of negatives.

\begin{figure*}[t]
  \centering
  \includegraphics[width=\linewidth]{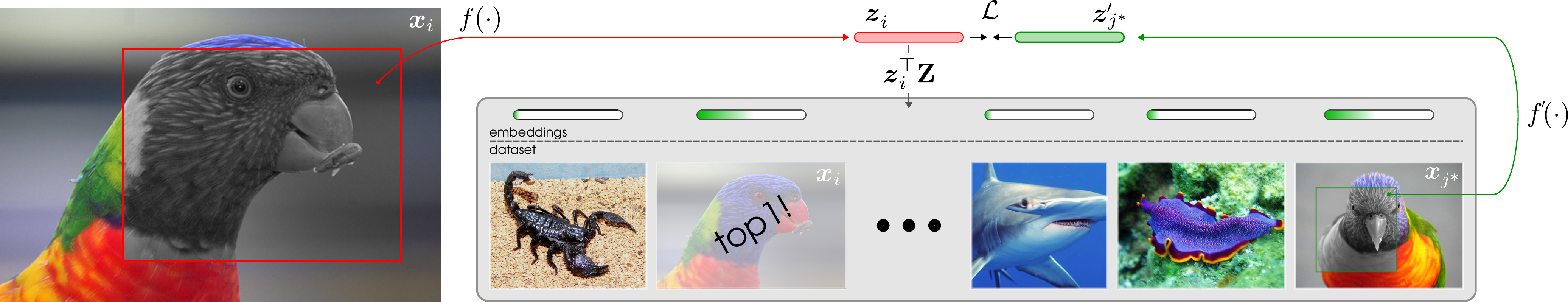}

  %\caption{\textbf{Overview of AdaSim.} Given an input image $\im_i$, we obtain two latent representations $\glob_i = f(t(\im_i))$ and $\glob_i^{'} = f'(t'(\im_i))$. Additionally, we sample another image $\im_{j^\star}$ in the dataset from $p^{win}(\im_j|\im_i)$ (see \cref{eq:similarity_distribution_windowed} and \cref{eq:p_im_dataset_windowed}) and obtain its latent representation $\glob_{j^\star}^{'} = f'(t'(\im_{j^\star}))$. We adaptively enforce a self-distillation loss $L$ between $\glob_i$ and $\glob_i^{'}$ or between $\glob_i$ and $\glob_{j^\star}^{'}$ here illustrated with a switch. In practice, only one of $\glob_i^{'}$ or $\glob_{j^\star}^{'}$ will be computed, see \cref{sec:adaptive_similarity_bootstrapping} and \cref{alg:adasim} for more details.}
  \caption{\textbf{Overview of AdaSim.} Given an input image $\im_i$, we obtain the latent representation $\glob_i = f(t(\im_i))$. Additionally, we sample another image $\im_{j^\star}$ in the dataset from $p^{win}(\im_j|\im_i)$ (see \cref{eq:similarity_distribution_windowed} and \cref{eq:p_im_dataset_windowed}) and obtain its latent representation $\glob_{j^\star}^{'} = f'(t'(\im_{j^\star}))$. A self-distillation loss $\mathcal{L}$ is enforced between $\glob_i$ and $\glob_{j^\star}^{'}$. For the sake of simplicity, only the scenario using bootstrapping is illustrated (see~\cref{alg:adasim}). Data augmentations are represented with grayscale bounding boxes.}
  \label{fig:main_figure}
\end{figure*}

Unfortunately, this combination does not work out of the box. \textbf{We empirically observe that it can be hurtful and even lead to collapse.} We scrutinize the reason for this unexpected behavior and provide a solution. We propose to estimate the quality of the latent space and adaptively use positive pairs sampled from a ranked set of neighbors (\cref{fig:query_support}) if the estimated quality of the latent space is high enough. This leads to an \textbf{Ada}ptive learning algorithm based on \textbf{Sim}ilarity bootstrapping dubbed AdaSim. The overall
framework is shown in \Cref{fig:main_figure}. We summarize our contributions as follows:
\begin{enumerate}
    \item We provide empirical evidence that when combined with self-distillation, straightforward bootstrapping as in \cite{dwibedi_little_2021} can lead to a performance drop or even collapse. This is validated for multiple self-distillation methods and backbone combinations;
    \item We propose an adaptive similarity bootstrapping learning method (AdaSim) in which the amount of bootstrapping is modulated via a single temperature parameter. Using a temperature parameter of 0, AdaSim defaults to self-distillation with standard positive image pairs generated from augmented views of the same image. We show that AdaSim performs best with a non-zero temperature parameter and outperforms the baselines on standard downstream tasks.
    % \item This is achieved with negligible compute and memory overhead.
    % \item \TODO{an image $\im_i$ and $\im_j$ should be considered as semantically close, not only if $f(t(\im_i))$ is close to $f(t(\im_i))$, but if $\mathbb{E}_{t \sim \mathcal{T}}[f(t(\im_i))]$ is close to $\mathbb{E}_{t' \sim \mathcal{T}}[f(t'(\im_j))]$?}
    % \item \Behzad{I would rather reduce the number of bullet points by merging some items, and add the last one referring to the superior performance of method over competing baselines on datasets, also, e.g., source code and models will be available upon acceptance (optional)}
\end{enumerate}

\section{Related work}
\label{sec:related_work}
%ithout access to labels, there does not exist a trivial objective to optimize. SSL objectives should be designed such that the learned features satisfy desirable properties. Desirable properties of features are akin to features with which solving a pretext task is easy. 
\noindent
{\bf Cross-view consistency} Early self-supervised methods make use of pretext tasks such as solving jigsaw puzzles~\cite{noroozi_jigsaw}, image rotation prediction~\cite{gidaris_rotations} and more \cite{contex_pred_doersch,contex_pred_Mundhenk,inpainting,predicting_noise}. Recently, there has been a shift towards learning features that are invariant to semantic preserving data augmentations \cite{dino,simsiam,moco,simclrv2,mocov2,mocov3,moby}. These data augmentations include geometric transforms (\eg \texttt{CROP}, \texttt{RESIZE} and \texttt{HORIZONTAL\_FLIP}) and photometric transforms (\eg \texttt{COLOR\_JITTER}, \texttt{SOLARIZE}, \texttt{GAUSSIAN\_BLUR} and \texttt{GRAYSCALE}). Stronger semantic preserving data augmentations lead to better downstream performance. However, the above-mentioned transforms lose their semantic preserving nature when they are too strong, \eg a very small \texttt{CROP} does not capture the object or a strong \texttt{GAUSSIAN\_BLUR} leads to a uniform image. %In the following subsections, we will \textcolor{red}{propose a way to generate strong semantic preserving augmentations.}

\noindent
{\bf Dense Cross-view consistency} Instead of applying coherence at the global-level, a more granular self-supervision can be obtained by enforcing cross-view consistency between matching local regions \cite{henaff2021efficient,Lebailly_2023_WACV,o2020unsupervised,stegmuller2023croc,wen2022slotcon,wang2021dense}.

\noindent
{\bf Neighbor bootstrapping} In order to generate strong semantic positive pairs less reliant on heuristics, NNCLR \cite{dwibedi_little_2021} proposes to use positive pairs of different images by bootstrapping nearest neighbors in the latent space. We describe their method in detail
 in \Cref{sec:bootstrapping_neighbors} as well as the issues that arise when used in conjunction with a self-distillation objective, which we try to overcome using adaptivity in \Cref{sec:adaptive_similarity_bootstrapping}. Similarly, \cite{Koohpayegani_2021_ICCV} proposes to bootstrap multiple neighbors for a single query.

\noindent
{\bf Clustering methods} Clustering methods \cite{asano2019self,caron2020unsupervised,deepcluster,DBLP:journals/corr/abs-1905-01278,clusterfit} also process multiple different images but do not make use of positive/negative pairs. They enforce structure in the latent space by learning prototypes and enforcing clusters to be compact.

\noindent
{\bf Queues/memory banks} Memory banks have mostly been used in the context of contrastive learning for storing negatives \cite{moco, mocov2, mocov3} reducing the need for (very) large batch sizes. \cite{dwibedi_little_2021} uses memory banks for mining positives while \cite{DBLP:journals/corr/abs-2112-01390} makes use of memory banks for mining both positives and negatives.

\section{Method}
\label{sec:method}

\subsection{Self-distillation vs contrastive learning}
\label{sec:selfdistillation-contrastive}
% old version
% Self-distillation and contrastive learning are the two of the most common learning scheme within self-supervised learning. Both schemes aim to learn features which are invariant to data-augmentations. This is done by enforcing similarity constraints between two augmentations of the same input image. They are similar in essence but differ in the way they avoid trivial solutions. Assume we dispose over a dataset $\mathcal{X}$ and an encoder $f$ from which we obtain a latent representation $\glob \in \mathcal{Z}$ of an image $\im \in \mathcal{X}$ via a forward pass of the image \ie $\glob = f(\im)$. Moreover, assume we dispose over an oracle $\mathcal{N}^+ \colon \mathcal{X} \to \mathcal{X}$ indicating valid positive pairs of images $(\im, \im^+)$ with $\im^+ \in \mathcal{N}^+(\im)$, an oracle $\mathcal{N}^- \colon \mathcal{X} \to \mathcal{X}$ indicating valid negative pairs of images $(\im, \im^-)$ with $\im^- \in \mathcal{N}^-(\im)$ and a distance metric\footnote{This is an abuse of terminology as $d$ does not necessarily have to satisfy all properties of a mathematical distance.} $d(\cdot \; , \; \cdot)$ defined in the latent space $\mathcal{Z}$. Valid positive pairs are image with the same semantic content and valid negative pairs are image with no shared semantic content. 

%a dataset $\mathcal{D} \subset \mathcal{X}$ and an

Self-distillation and contrastive learning are ubiquitous within self-supervised learning. Both schemes aim to learn discriminative features in the absence of labels. This is mainly done by enforcing similarity constraints between two augmentations of the same input image. The two methods are similar in essence but differ in the way they avoid trivial solutions. Assume we dispose of an encoder $f$ from which we obtain a latent representation $\glob \in \mathcal{Z}$ of an image $\im \in \mathcal{X}$, \ie $\glob = f(\im)$ with $\mathcal{Z}$ and $\mathcal{X}$ being a latent- and image space, respectively. Moreover, assume we dispose of an oracle $\mathcal{N}^+$ indicating valid positive pairs of images $(\im, \im^+) \in \mathcal{N}^+$, an oracle $\mathcal{N}^-$ indicating valid negative pairs of images $(\im, \im^-) \in \mathcal{N}^-$ and a distance metric\footnote{This is an abuse of terminology as $d$ does not necessarily have to satisfy all properties of a mathematical distance.} $d(\cdot \; , \; \cdot)$ defined in the latent space $\mathcal{Z}$. Valid positive pairs are images with the same semantic content and valid negative pairs are images with no shared semantic content.

% \begin{equation}
%     \mathcal{L}_{contra} = F(d(f(\im), f(\im^+)), \{d(f(\im), f(\im^-)) \colon \im^- \in \mathcal{N}^-\})
% \end{equation}

%A contrastive loss (that gets minimized) is a decreasing function w.r.t. the positive terms and an increasing function w.r.t. the negative terms.
\noindent
{\bf Contrastive objective}
% \subsection{Contrastive objective}
A contrastive learning loss relies on attraction and repelling mechanisms: the former enforces similarity between positive pairs and the latter enforces dissimilarity between the negative pairs. Formally, the attraction term is of the form $d(f(\im), f(\im^+))$ and the repelling terms are of the form  $d(f(\im), f(\im^-))$. Here, we refer to ``term'' in its broad sense and therefore do not necessarily refer to an additive term. Usually, there are many negative terms for a single positive term. One famous example of such contrastive loss is the InfoNCE loss \cite{sohn_improved_2016,Contrastive_Predictive_Coding,DBLP:journals/corr/abs-1805-01978} defined as:
% \begin{equation}
%     \mathcal{L}_{\text{contra}} = -\log\left(\frac{\exp(f(\im)^\top f(\im^+))}{\exp(f(\im)^\top f(\im^+)) + \sum\limits_{\im^-\in \mathcal{S}^{\text{neg}}}{\exp(f(\im)^\top f(\im^-))}}\right)
%     \label{eq:info-nce}
% \end{equation}

\begin{equation}
    \mathcal{L}_{\text{contra}} = -\log\left(\frac{\exp(s^+/\tau)}{\exp(s^+/\tau) + \sum\limits_{s^-}{\exp(s^-/\tau)}}\right)
    \label{eq:info-nce}
\end{equation}
where $s^{+}=f(\im)^\top f(\im^+)$ and $s^{-}=f(\im)^\top f(\im^-)$. $(\im,\im^+)$ is sampled from $\mathcal{N}^+$ and $(\im,\im^-)$ are sampled from $\mathcal{N}^-$. The distance metric $d(\cdot \; , \; \cdot)$ is defined as the scalar product $\langle \cdot \; , \; \cdot \rangle$. The total contrastive objective is \Cref{eq:info-nce} summed over all training images $\im$.

\noindent
{\bf Self-distillation objective}
% \subsection{Self-distillation objective}
As opposed to the contrastive scenario, the self-distillation objective does not use negative image pairs to avoid the collapse to trivial solutions but uses asymmetry between the two branches. The form the asymmetry takes (momentum encoder, additional predictor on one branch, using stop-gradients in one branch \etc \cite{dino,simsiam,byol}) can be abstracted out. Given two encoders $f$ and $f'$, a self-distillation loss only has positive terms of the form% $d(f(\im), f'(\im^+))$. 

% In SimSiam \cite{simsiam}, $f'(\im) := h(f(\im))$ with $h \colon \mathcal{Z} \to \mathcal{Z}$ a predictor and $d$ is defined as:
% \begin{equation}
%     d(f(\im), f'(\im^+)) := -\frac{f(\im)^\top f'(\im^+)}{\norm{f(\im)}_2 \norm{f'(\im^+)}_2}
% \end{equation}

% In DINO \cite{dino}, $f'$ is a momentum encoder $f'(\im) := f_{\text{EMA}}(\im)$ whose weights are updated at the end of each epoch as follows: $\weight' \leftarrow \lambda \weight' + (1-\lambda) \weight$. $\weight$ and $\weight'$ are the weights of $f$ and $f'$, respectively. Given a head $h \colon \mathcal{Z} \to \mathcal{P}$ mapping the latent space discrete probability mass functions, and its exponential moving average equivalent $h'$, $d$ is defined as:

% \begin{equation}
%     d(f(\im), f'(\im^+)) := H(h(f(\im)), h'(f'(\im^+)))
% \end{equation}
% where $H$ is the cross entropy
% \begin{equation}
%     H(p, q) = -\sum_{i \in \mathcal{I}} p(i) \log q(i)
% \end{equation}
% and $\mathcal{I}$ is the support of the distributions $p$ and $q$, \ie $\mathcal{I}=[I]=\{1, 2, \cdots, I\}$.

% In self-distillation methods (\eg DINO and SimSiam), since there are no negative terms, the self-distillation objective is simply:
\begin{equation}
    \mathcal{L}_{distil} = d(f(\im), f'(\im^+))
    \label{eq:self-distil}
\end{equation}
for a given positive pair $(\im, \im^+)$. The total self-distillation objective is \Cref{eq:self-distil} summed over all positive pairs $(\im, \im^+) \in \mathcal{N}^+$.

\subsection{Bootstrapping neighbors in the latent space}
\label{sec:bootstrapping_neighbors}
In the absence of oracle $\mathcal{N}^+$ and $\mathcal{N}^-$, most (if not all) previous work approximate $\mathcal{N}^-$ with random image pairs. Given a distribution $\mathcal{T}$ of semantic preserving data augmentations, $\mathcal{N}^+$ is usually approximated with pairs of random augmentations from the same input image, \ie $(t(\im), t'(\im))$ where $t$ and $t'$ are sampled from $\mathcal{T}$. The stronger the semantic preserving augmentations $t$ and $t'$ are, the better the learned features become. However, their semantic preserving nature will be lost if they are made too strong. 

To obtain more complex and diverse pairs of positive images, NNCLR \cite{dwibedi_little_2021} proposes to approximate $\mathcal{N}^+$ with pairs of nearest neighbors. More precisely, given two latent representations ($\glob$ and $\glob'$) of the same image $\im$ and a FIFO queue $Q$ of previously computed representations (with $|Q| < |\mathcal{D}|$), positive pairs are defined as $(\glob', \text{NN}(\glob, Q))$, where the nearest neighbor operator is defined as:
\begin{equation}
    \text{NN}(\glob, Q) = \argmin_{q \in Q} \norm{z-q}_2
\end{equation}
Note here that the positive pairs are defined in the latent space $\mathcal{Z}$ and not in the image space. Under the assumption that the latent space properly captures the semantics of images, these pairs of neighbors are expected to share the same semantic content but their representation may still be slightly different. Enforcing similarity constraints between the two representations would help to learn features that are invariant to everything but the semantics of the image (\eg class label information). However, two issues arise when relying exclusively on nearest neighbors as positive pairs:
\begin{enumerate}
    \item \label{iss:one} Using only neighbors as positive pairs, \ie not relying on augmented views as positive pairs, leaves out valuable self-supervisory signal. Using standard positive pairs of augmented views from the same image is desirable to explicitly learn data-augmentation invariant features, but that is not enforced.
    \item \label{iss:two} The latent space might not capture the semantics of the image well, \ie the positive pair is wrong. This would lead to undesirable gradient flows, \eg pulling an image of a cat closer to an image of a building.
\end{enumerate}

Throughout the paper, we refer to the above as issue \ref{iss:one} and issue \ref{iss:two}. Using a contrastive objective, the impact of these issues is limited since informative gradient signal can still be obtained from the negative pairs which in practice are almost always correct (random). Using a self-distillation objective, we empirically observe that the above issues are problematic to the point that the downstream performance can be worse than using standard positive pairs using data-augmentations (\cref{sec:results}, \cref{tab:main_table}, \cref{tab:few_shot}).

\subsection{Adaptive similarity bootstrapping}
\label{sec:adaptive_similarity_bootstrapping}
\subsubsection{Need for standard positive pairs}
\label{sec:need_for_standard_positive_pairs}

To avoid issue \ref{iss:one}, we adaptively use augmentations of the same image or of a neighbor to form a positive pair. To do this, we propose to work with a cache that has the same size as the dataset $\mathcal{D}$ as opposed to using the queue $Q$ from NNCLR \cite{dwibedi_little_2021}. Using a small queue, it is very unlikely to encounter a representation originating from the same image. We denote the cache with $\Z \in \mathbb{R}^{N \times d}$, where $N$ is the size of the dataset and $d$ is the dimension of the latent space. At the end of the forward pass, the current latent representation $\glob_i$ of an augmentation of the $i$-th image $\im_i \in \mathcal{D}$ (\ie $f(t(\im_i))$ with $t \sim \mathcal{T}$) is updated in the cache. As such, $\Z$ holds a latent representation for every image in the dataset at all times. Given a latent representation $\glob_i$ of the $i$-th image and the cache $\Z$, we can define a similarity metric $m_i(j)$ between image $i$ and all images $\im_j$:
\begin{equation}
    m_i(j) = \glob_i^\top \Z_j
\end{equation}
where $\Z_j$ refers to the latent representation of image $j$ in the cache. $m_i(j)$ can in turn be mapped into a similarity distribution using a softmax normalization:

\begin{equation}
    s_i(j) = \frac{\exp{\left(m_i(j)\right / \tau)}}{\sum_{k\in [|\mathcal{D}|]} \exp{\left(m_i(k) / \tau\right)}}, \quad \quad i,j \in [|\mathcal{D}|]
    \label{eq:similarity_distribution}
\end{equation}
where $\tau$ is a temperature parameter modulating the sharpness of the distribution (ablation in \cref{tab:ablation}). We can now define an isomorphic probability distribution over the images in the dataset:

\begin{equation}
    p(\im_j | \im_i) = s_i(j), \quad \quad \im_i,\im_j \in \mathcal{D}%^2
    \label{eq:p_im_dataset}
\end{equation}

To approximate the oracle $\mathcal{N}^+$ of positive pairs, we propose to use image $i$ and an image sampled from the similarity distribution. That is, we form positive pairs of the form $(t(\im_i), t'(\im_{j^\star}))$ with $\im_{j^\star}$ sampled from $p(\im_j | \im_i)$ and with $t$ and $t'$ sampled from $\mathcal{T}$. Note that $\Z$ contains features for all images, not excluding image $\im_i$. Therefore, we always have a non-zero probability of having a positive pair generated from the same input image which mitigates issue \ref{iss:one} from \Cref{sec:bootstrapping_neighbors}. Sampling positive pairs of the form $(t(\im_i), t'(\im_{j^\star}))$ also allows for the possibility to sample more diverse and complex pairs of positives compared to the case when we only consider top-1 neighbors, as can be seen in \Cref{fig:query_support}. This diversity can be increased by increasing the temperature $\tau$.

% The probability to sample a positive pair of the form $(t(\im), t'(\im))$ can be negligibly low. To solve this, we impose that $()$

\subsubsection{Need for adaptivity}
\label{sec:need_for_adaptivity}
Recall that issue \ref{iss:two} from \Cref{sec:bootstrapping_neighbors} is that the latent space might not capture the semantics of images properly (especially at the beginning of the pretraining). That is, neighbors in the latent space might have completely unrelated semantic content. We propose to estimate the quality of the latent space by observing how close two different augmentations of the same input image $\im_i$ are mapped via the encoder $f$. If this distance is low compared to that of the latent representations of other images in the cache $\Z$, then it means that the encoder $f$ is good at mapping images similar to $\im_i$ close together. In that case, we can expect the vicinity of the queried image $\im_i$ to also share semantic content with image $\im_i$ and can therefore use elements of the vicinity to form a positive pair with $\im_i$. If this distance is too high, we default to a standard positive pair composed of two augmentations of the same input image. Mathematically, if $\argmax_{\im_j} p(\im_j | \im_i) == \im_i$, then we sample $\im_{j^\star}$ from $p(\im_j | \im_i)$ and use a positive pair $(t(\im_i), t'(\im_{j^\star}))$ with $t$ and $t'$ sampled from $\mathcal{T}$. Otherwise, we use a standard positive pair $(t(\im_i), t'(\im_i))$.

% \TODO{mention something about this decision being instance specific.}

\subsubsection{How to rank neighbors?}
We propose to extend the adaptive framework to account for the similarity history over the past epochs. The rationale behind this is that the similarity between two images $t(\im_i)$ and $t'(\im_j)$ can be strongly affected by $t$ and $t'$, especially at the beginning of the pretraining. For example, given a randomly initialized encoder $f$, the similarity between $f(t(\im_i))$ and $f(t'(\im_j))$ will be mostly determined by how similar $t$ and $t'$ are. Therefore, an image $\im_i$ and $\im_j$ should be considered as semantically close, not only if $f(t(\im_i))$ is close to $f(t'(\im_j))$, but if $\mathbb{E}_{t \sim \mathcal{T}}[f(t(\im_i))]$ is close to $\mathbb{E}_{t' \sim \mathcal{T}}[f(t'(\im_j))]$.

In practice, we do not have access to the true expectation and therefore take the empirical mean over the last $w$ epochs. More precisely, we define the similarity metric for a given epoch $e$ which we denote with the superscript $^{(e)}$:
\begin{equation}
    m_i^{(e)}(j) = \left(\glob_i^\top \Z_j\right)^{(e)}
    \label{eq:m_i_j^e}
\end{equation}
and average this similarity metric over the last $w$ epochs to obtain a windowed similarity metric for the current epoch $E$:

\begin{equation}
    m_i^\text{win}(j) = \frac{1}{w}\sum\limits_{e \in \mathcal{W}_E^w}m_i^{(e)}(j)
    \label{eq:windowed_sim_metric}
\end{equation}
where $\mathcal{W}_E^w = \{E-w+1, E-w+2, \cdots E\}$ denotes the set of the previous $w$ epochs with epoch $E$ being the current epoch. Similarly to \Cref{eq:similarity_distribution}, we can define:

\begin{equation}
    s_i^\text{win}(j) = \frac{\exp{\left(m_i^\text{win}(j)\right / \tau)}}{\sum_{k\in [|\mathcal{D}|]} \exp{\left(m_i^\text{win}(k) / \tau\right)}}, \quad \quad i,j \in [|\mathcal{D}|]
    \label{eq:similarity_distribution_windowed}
\end{equation}
where $\tau$ is a temperature parameter as in \Cref{eq:similarity_distribution}. And similarly to \Cref{eq:p_im_dataset}, we can define:
\begin{equation}
    p^{\text{win}}(\im_j | \im_i) = s_i^\text{win}(j), \quad \quad \im_i,\im_j \in \mathcal{D}%^2
    \label{eq:p_im_dataset_windowed}
\end{equation}

From here on, we use $p^{\text{win}}(\im_j | \im_i)$ instead of $p(\im_j | \im_i)$ as the sampling distribution. At the beginning of the pretraining, \ie as long as no $w$ similarity metrics have been computed yet, we default to using standard positive pairs generated from augmented views of the same image. Note that for a window of size 1 ($w=1$), we fall back to \Cref{eq:similarity_distribution} and \Cref{eq:p_im_dataset} from \Cref{sec:adaptive_similarity_bootstrapping}, \ie $s_i(j) = s_i^\text{win}(j)$ and $p(\im_j | \im_i)=p^{\text{win}}(\im_j | \im_i)$.

\begin{algorithm}[t]
	\caption{\texttt{AdaSim}: Adaptive Similarity Bootstrapping framework} 
    \textbf{Input:} $\mathcal{D}$: an unlabeled dataset, $\mathcal{T}$: a distribution over the possible augmentations, $f$: an encoder parametrized with weights $\weight$, \texttt{OPTIMIZER}: an optimizer, $\Z \in \mathbb{R}^{N \times d}$: a zero-initialized cache ($N=|\mathcal{D}|$ and $d$ is the dimension of the latent space), $w$: window size, $L$: self-distillation loss\\
    \textbf{Output:} Trained weights 
	\begin{algorithmic}[1]
	\For{$e \in \{1, 2, \cdots \text{NB\_EPOCHS}\}$}
		
		\For {$i \in [|\mathcal{D}|]$}
		    \State Sample $t$ and $t'$ from $\mathcal{T}$
		  %  \State $\aug_{i} = t(\im_i)$
		    \State $\rep_{i} = f(t(\im_i))$
		    \State $m_i^{(e)}(j) = \left(\glob_i^\top \Z_j\right)^{(e)}$ \Comment{\cref{eq:m_i_j^e}}
		    \State \texttt{update}$(\Z, \rep_{i})$ \Comment{Update cache with $\rep_i$}
		    
    	    \If{$e \leq w$}
    		    \State $\mathcal{L} = L(t(\im_i), t'(\im_i))$
    	    \Else 
    	        \State $m_i^{\text{win}}(j) = \frac{1}{w}\sum\limits_{e' \in \mathcal{W}_e^w}m_i^{(e')}(j)$ \Comment{\cref{eq:windowed_sim_metric}}
    	        \State $s_i^{\text{win}}(j) = \frac{\exp{\left(m_i^{\text{win}}(j) / \tau\right)}}{\sum_k \exp{\left(m_i^{\text{win}}(k) / \tau\right)}}$  \Comment{\cref{eq:similarity_distribution_windowed}}
    	        \State $p^{\text{win}}(\im_j | \im_i) = s_i^{\text{win}}(j)$ \Comment{\cref{eq:p_im_dataset_windowed}}
		  %  \If{$i == \argmax\limits_{j \in \bigcup\limits_{e\in\mathcal{W}_E^w} \mathcal{S}^{(e)}}{s^{\text{win}}(j)}$}
		    \If{$\im_i == \argmax\limits_{\im_j}{p^{\text{win}}(\im_j | \im_i)}$}  
		        \State Sample $\im_{j^\star}$ from $p^{\text{win}}$
          \State $\mathcal{L} = L(t(\im_i), t'({\im_{j^\star}}))$ \hspace*{3.5em} \rlap{\smash{$\left.\begin{array}{@{}c@{}}\\{}\\{}\\{}\\{}\\{}\end{array}\right\}%
          \begin{tabular}{c}Adaptive\\sampling\end{tabular}$}}
		    \Else
		        \State $\mathcal{L} = L(t(\im_i), t'(\im_i))$
		    \EndIf
    		  %  \State Sample $t'$ from $\mathcal{T}$
    		  %  \State $\mathcal{L} = L(t(\im_i), t'({\im_{j^\star}}))$
    	    \EndIf
		    \State $\weight \leftarrow$ \texttt{OPTIMIZER}$(\weight, \nabla_{\weight}\mathcal{L})$
		    
		\EndFor
% 		\State $\weight_t \leftarrow \lambda \weight_t + (1-\lambda) \weight_s$
	\EndFor
	\State \Return $\weight$
	\end{algorithmic}
	\label{alg:adasim}
\end{algorithm}

\subsection{Memory and compute overhead}
\noindent
{\bf Memory overhead} In practice storing $w$ versions of $m_i^{(e)}(j)$ with $e \in \mathcal{W}_E^w$ is not feasible when the dataset is large as it would require storing $w$ entries for each pair of images. In the case of ImageNet-1k \cite{imagenet}, that would require about $(1.3\text{M})^2 \times 4 \times w \sim 300$TB which is infeasible\footnote{$1.3\text{M}$ refers to the size of the dataset and 4 bytes are required to store a single float entry of 32 bits.}. However, since we sample from the similarity distribution to form positive pairs, we are only interested in the most similar images. Therefore, we can restrict the support of $m_i^{(e)}(j)$ to the $K$ highest elements. We denote this new support as $\mathcal{S}^{(e)}$ with $|\mathcal{S}^{(e)}|=K$. Note that for every epoch $e$, the support of $m_i^{(e)}(j)$ is different. The similarity metric $m_i^\text{win}(j)$ from \Cref{eq:windowed_sim_metric} with restricted domain is obtained as follows:

\begin{equation}
    m_i^\text{win}(j) = \frac{1}{w}\sum\limits_{e \in \mathcal{W}_E^w}\mathbbm{1}_{\{j\in \mathcal{S}^{(e)}\}}m_i^{(e)}(j)
\end{equation}
where $\mathbbm{1}$ denotes the indicator function and with $j \in \mathcal{S}_{\text{union}}$ and  $\mathcal{S}_{\text{union}} = \bigcup_{e\in\mathcal{W}_E^w} \mathcal{S}^{(e)}$. The only difference for the similarity distribution from \Cref{eq:similarity_distribution_windowed} is that its support is limited to $\mathcal{S}_{\text{union}}$. Similarly for \Cref{eq:p_im_dataset_windowed}, the only difference is that its support is limited to $\mathcal{S}_{\text{win}} = \{\im_j : j \in \mathcal{S}_{\text{union}}\}$. Taking all the above into consideration, the final algorithm AdaSim is illustrated in \Cref{alg:adasim}.

\noindent
{\bf Compute overhead} The compute overhead is limited to the projection of a representation $\glob$ onto the cache $\Z$ which is embarrassingly parallelizable on GPU. This requires about $d|\mathcal{D}|=0.5$B operations (for ViT-S/16) which is much less than the 4.6B FLOPs in the backbone (see \cref{sec:runtime}).

\section{Results}
\label{sec:results}

\subsection{Rationale of the experiment design}
The goal of the paper is \textbf{1)} to show that bootstrapping neighbors using a self-distillation objective can hinder the performance or \textbf{2)} even lead to collapse and \textbf{3)} ultimately propose an adaptive bootstrapping scheme which not only solves the above-mentioned issues but also improves on the baselines using standard positive pairs. To achieve this goal, we compare two self-distillation methods (SimSiam \cite{simsiam} and DINO \cite{dino}) with different backbones (ViT-S/16 \cite{vit16x16} and ResNet-50 \cite{resnet}) in a simple controlled setup (pretraining on ImageNet-1k \cite{imagenet}, same hyperparameters, using only 2 global crops). For every evaluation, we compare 1) the baseline with 2) the baseline + straightforward nearest neighbor bootstrapping \cite{dwibedi_little_2021} and 3) the baseline + AdaSim.

\begin{table}[t]
\caption{\textbf{Supervised oracle.} $p$ indicates the probability to sample a standard positive pair ($1-p$ is the probability to sample a supervised positive pair, see \cref{sec:supervised_oracle}).}
\resizebox{\columnwidth}{!}{
\begin{tabular}{c|cccc}
\toprule
$p$ & $k$-NN (top-1) & $k$-NN (top-5) & linear (top-1) & linear (top-5) \\
\midrule
0                            & 74.3                            & 90.5                            & 75.8                               & 92.7                               \\
0.5                          & 74.9                            & 90.9                            & 76.3                               & 93.0                              \\
\bottomrule
\end{tabular}
}
\label{tab:supervised}
\end{table}

We report results on the linear and $k$-NN benchmarks of ImageNet-1k which are industry standard evaluation protocols for self-supervised methods (\cref{sec:imagenet-benchmark}). To evaluate how generalizable the learned features are, we further compare all methods on few-shot transfer downstream tasks (\cref{sec:fewshot}). Then, we run an ablation study on AdaSim-specific hyperparameters (\cref{sec:ablations}) and finish with some interesting training metrics that are helpful to understand AdaSim intuitively (\cref{sec:underthehood}). The main takeaway from this section is that AdaSim avoids issues \ref{iss:one} and \ref{iss:two} incurred by straightforward nearest neighbor bootstrapping and shows performance improvements on all downstream tasks.

\begin{table}[t]
\definecolor{darkgreen}{rgb}{0, 0.5, 0}
\centering
% \caption{\textbf{Comparison of different methods on the $k$-NN and linear evaluation benchmark on ImageNet-1k \cite{imagenet}.} The 4 blocks of rows correspond to 4 different baselines: SimSiam ResNet-50, DINO-2 ResNet-50, DINO-2 ViT-S/16 and DINO-2 ViT-B/16. Each baseline is compared against straightforward nearest neighbor (NN) bootstrapping and against AdaSim. ``\collapse'' denotes that the training objective does not converge, \eg due to collapse. Rows corresponding to AdaSim are \hl{highlighted}. \textbf{Bold} text is used for the best-performing row within each block.}
\caption{
\textbf{Linear evaluation and $k$-NN benchmarks on ImageNet-1k \cite{imagenet}.} We report the performance of the proposed bootstrapping scheme in conjunction with various self-distillation methods and backbones. \hl{AdaSim} is systematically compared against the settings where no bootstrapping occurs and the one using straightforward bootstrapping (+NN).
``\collapse'' denotes a failure to converge.}
\resizebox{\columnwidth}{!}{
\begin{tabular}{lcccc}
\toprule
Method           & Model    & Epochs & $k$-NN & Linear \\
\midrule
SimSiam \cite{simsiam}         & ResNet-50 & 100   &   57.1   &    68.0    \\
SimSiam + NN     & ResNet-50 & 100   &   56.2 \textcolor{red}{(- 0.9)}  &    65.9 \textcolor{red}{(- 2.1)}  \\
\rowcolor{cyan!20} SimSiam + AdaSim & ResNet-50 & 100   &   \textbf{57.9} \textcolor{darkgreen}{(+ 0.8)}  &    \textbf{68.1} \textcolor{darkgreen}{(+ 0.1)}   \\
\midrule
DINO-2  \cite{dino}         & ResNet-50 & 100   &   50.2   &   60.0     \\
DINO-2 + NN      & ResNet-50 & 100   &    \collapse  &   \collapse     \\
\rowcolor{cyan!20} DINO-2 + AdaSim  & ResNet-50 & 100   &  \textbf{50.7} \textcolor{darkgreen}{(+ 0.5)}   &    \textbf{60.1} \textcolor{darkgreen}{(+ 0.1)}   \\
\midrule
DINO-2   \cite{dino}        & ViT-S/16  & 800   &  68.4    &   71.9     \\
DINO-2 + NN     & ViT-S/16  & 800   &   \collapse   &    \collapse    \\
% \rowcolor{cyan!10} DINO-2 + AdaSim  & ViT-S/16  & 800   &  \textbf{69.3}    &     \textbf{72.4}  \\
% \midrule
\rowcolor{cyan!20} DINO-2 + AdaSim  & ViT-S/16  & 800   &  \textbf{70.1} \textcolor{darkgreen}{(+ 1.7)}    &     \textbf{73.3} \textcolor{darkgreen}{(+ 1.4)}  \\
\midrule
DINO-2   & ViT-B/16    & 800 & 69.2 & 73.5  \\
DINO-2 + NN & ViT-B/16 & 800 & \collapse & \collapse \\
\rowcolor{cyan!20} DINO-2 + AdaSim & ViT-B/16 & 800 &  \textbf{72.7} \textcolor{darkgreen}{(+ 3.5)} & \textbf{75.0} \textcolor{darkgreen}{(+ 1.5)} \\
\bottomrule
\end{tabular}
}
\label{tab:main_table}
\end{table}

\begin{table*}[t]
\centering
\caption{\textbf{Few-shot transfer (5-way 5-shot) using prototypical networks \cite{snell_prototypical_2017} on multiple standard datasets.} The reported metrics are top-1 accuracy for Food, SUN397, Cars, DTD and mean per-class accuracy for the other datasets. ``\collapse'' denotes that the training objective does not converge, \eg due to collapse. Rows corresponding to AdaSim are \hl{highlighted}. \textbf{Bold} text is used for the best performing row within each block.}
\footnotesize
\resizebox{\textwidth}{!}{
\begin{tabular}{lcc|ccccccccc}
\toprule
Method & 
Model & 
Epochs & 
Aircraft \cite{aircraft_dataset} & 
Caltech101 \cite{caltech101} & 
Cars \cite{krause_collecting_nodate} & 
DTD \cite{dtd_dataset} & 
Flowers \cite{flower_dataset} & 
Food \cite{food101} & 
Pets \cite{pets_dataset} & 
SUN397 \cite{sun397} & 
Avg   \\
\midrule
SimSiam \cite{simsiam}         & Resnet-50 & 100                        & 44.13    & 94.88      & 51.46 & \textbf{78.94} & 94.19   & 68.12 & 88.27 & \textbf{91.12}  & 76.39 \\
SimSiam + NN   & Resnet-50 & 100                        & 43.44    & 94.4       & 50.52 & 76.69 & 93.85   & 67.04 & 88.66 & 90.41  & 75.63 \textcolor{red}{(- 0.76)}\\
\rowcolor{cyan!20} SimSiam + AdaSim & Resnet-50 & 100                        & \textbf{45.71}    & \textbf{94.91}      & \textbf{51.71} & 78.87 & \textbf{94.54}   & \textbf{68.38} & \textbf{88.91} & 91.0   & \textbf{76.75} \textcolor{darkgreen}{(+ 0.36)}\\
\midrule

DINO-2 \cite{dino} & Resnet-50 & 100 & \textbf{40.19} & 92.65 & 45.84 & \textbf{79.58} & \textbf{90.00} & 63.22 & 80.35 & \textbf{90.58} & 72.80\\
DINO-2 + NN  & Resnet-50 & 100 & \collapse & \collapse & \collapse & \collapse & \collapse & \collapse & \collapse & \collapse & \collapse \\
\rowcolor{cyan!20} DINO-2 + AdaSim & Resnet-50 & 100 & 38.80 & \textbf{92.69} & \textbf{46.58} & 79.54 & 89.68 & \textbf{64.52} & \textbf{81.52} & 90.41 & \textbf{72.97} \textcolor{darkgreen}{(+ 0.17)}\\
\midrule

DINO-2 \cite{dino}   & ViT-S/16  & 800                        & 52.78    & 98.4       & 56.42 & 81.87 & 96.54   & 76.06 & 96.04 & 94.26  & 81.55 \\
DINO-2 + NN  & ViT-S/16 & 800 & \collapse & \collapse & \collapse & \collapse & \collapse & \collapse & \collapse & \collapse & \collapse \\
\rowcolor{cyan!20} DINO-2 + AdaSim          & ViT-S/16  & 800                        & \textbf{56.54} & \textbf{98.93} & \textbf{58.04} & \textbf{82.35} & \textbf{96.96} & \textbf{77.23} & \textbf{96.54} & \textbf{94.78} & \textbf{82.67} \textcolor{darkgreen}{(+ 1.12)} \\
\midrule
Supervised & Resnet-50 &  & 58.35 & 97.61 & 73.68 & 80.83 & 94.19 & 76.23 & 97.45 & 93.78 & 84.02\\
\bottomrule
\end{tabular}
}
\label{tab:few_shot}
\end{table*}%

\subsection{Evaluation benchmarks}
\noindent
{\bf Linear evaluation} A linear layer is stacked on top of the frozen features and trained on the training set of the downstream task. We report the top-1 accuracy on the test set. For each setting, we use the evaluation protocol (\eg choice of optimizer, number of training epochs \etc) from the corresponding baseline (SimSiam \cite{simsiam} or DINO \cite{dino}).
To evaluate the intrinsic quality of representations, the downstream evaluation should ideally not require many learnable parameters. In the case of ResNet-50, the number of parameters in the linear layer is $1000*d$ where $d=2048$ which is about $2$ million parameters. The following evaluations do not have any learnable parameters and are thus better suited to evaluate the intrinsic quality of the pretraining.

\noindent
{\bf $\boldsymbol{k}$-NN evaluation} The representation $\glob$ of each image in both the training and test set is computed. Then each image in the test set gets a label assigned based on votes from the nearest neighbors in the training set. We use $k=20$ to stay consistent with previous work and report the top-1 accuracy.

\noindent
{\bf Few-Shot transfer} This evaluation uses a nearest-centroid classifier (Prototypical Networks \cite{snell_prototypical_2017}). We use the code and datasets (except CIFAR-10 and CIFAR-100 because the images are only 32x32) from \cite{ericsson_how_2021}. We consider 5-way 5-shot transfer with a query set of 15 images and average results over 600 randomly sampled few-shot episodes.

% \begin{table*}[b]
%     \centering
%     \begin{tabular}{c|ccccc}
%         \toprule
%         \multicolumn{2}{c}{$\tau$}
%         \midrule
%         ImageNet $k$-NN top-1 & 68.4 & 68.4 & 68.9 & 69.3 & 68.6\\
%         ImageNet $k$-NN top-5 & 86.6 & 86.8 & 87.1 & 87.4 & 86.9\\
%         \bottomrule

%     \end{tabular}
%     \caption{Caption}
%     \label{tab:my_label}
% \end{table*}

\subsection{Implementation details}
\label{sec:implementation_details}
For both \href{https://github.com/facebookresearch/dino}{DINO} and \href{https://github.com/facebookresearch/simsiam}{SimSiam}, the same hyperparameters are used as reported on their GitHub. To make sure the size of the queue/cache does not impact the results, we implement the ``baseline + NN'' entries in \Cref{tab:main_table} and \Cref{tab:few_shot} with a cache that has the size of the whole dataset. To confirm the fact that standard positive pairs are needed (see issue \ref{iss:one}), we implement the querying of the nearest neighbor such that it cannot originate from the same image $\im_i \in \mathcal{D}$ (as is the case with a queue of small size). More implementation details can be found in \Cref{sec:implementation_details}.

\subsection{Supervised oracle}
\label{sec:supervised_oracle}
As a starter, to confirm our intuition that better positive pairs lead to better performance on downstream tasks, we approximate the oracle of positive pairs $\mathcal{N}^+$ using the labels from ImageNet-1k \cite{imagenet}. We sample a positive pair as two random images from the same class, on top of which we still apply augmentations. Empirically, we observe that the convergence (speed) is much worse than using standard positive pairs. To speed up the convergence, we sample standard positive pairs with a certain probability. Given that it makes sense for this probability to be high at the beginning of the pretraining, we simply try a linear schedule going from 1 to $p$. Results for $p=0$ and $p=0.5$ can be found in \Cref{tab:supervised}. It can be observed that $p=0.5$ performs better which corroborates our reasoning related to issue \ref{iss:one}.

\subsection{ImageNet-1k benchmarks}
\label{sec:imagenet-benchmark}
The $k$-NN and linear evaluation results on ImageNet-1k \cite{imagenet} are reported in \Cref{tab:main_table}. The last block in blue shows the best performing setting ($\tau=0.2$, $w=50$, $K=10$) from the ablation in \Cref{tab:ablation} with a long pretraining schedule of 800 epochs. The first 2 blocks of rows are trained with a window size of $w=1$ (and $\tau=0.2$, $K=10$) to confirm that AdaSim does not require a large window to improve the baseline and avoid collapse. DINO-2 denotes DINO with only 2 global crops. First, we can observe that with DINO-2 \cite{dino}, straightforward nearest neighbor bootstrapping (NN) does not converge (illustrated with ``\collapse''). This is confirmed for different backbones and training schedules. DINO-2 + AdaSim does converge and improves the baselines. The training objective of SimSiam \cite{simsiam} + NN does converge but suffers from a performance impact. %Note that the relative difference between different runs is higher for the $k$-NN benchmark as no parameters are learned.

%%%%%%%%%%% NEW TABLE %%%%%%%%%%%%%%
\begin{table*}[t]
\centering
\caption{
\textbf{Ablation study over AdaSim specific hyperparameters.} The ablation is run over 800 epochs. If not otherwise specified, the values of the hyperparameters are $(\tau, w, K) = (0.2, 10, 3)$.
}
\footnotesize
%\resizebox{1\textwidth}{!}{
\begin{tabular}{l c c c c c c c c c c c c c c c c}
\toprule
 && \multicolumn{5}{c}{Temperature ($\tau$)} && \multicolumn{3}{c}{Window size ($w$)} && \multicolumn{5}{c}{Support size ($K$)} \\
\cmidrule{3-7}  \cmidrule{9-11} \cmidrule{13-17}
 && 0.0 & 0.05 & 0.1 &  \cellcolor{cyan!20} 0.2 & 0.4 && 1 & \cellcolor{cyan!20} 10 & 50 && 2 & \cellcolor{cyan!20} 3 & 5 & 10 & 20 \\
\midrule
$k$-NN && 68.4 & 68.8 & 69.0 & \cellcolor{cyan!20} \textbf{70.1} & 69.8 && 69.3 & \cellcolor{cyan!20} \textbf{70.1} & 69.4 && 69.4 & \cellcolor{cyan!20}\textbf{70.1} & 70.1 & 69.1 & 68.4 \\
linear && 71.9 & 72.4 & 72.6 & \cellcolor{cyan!20}\textbf{73.3} & 73.2 && 72.6 & \cellcolor{cyan!20} \textbf{73.3} & 72.7 && 72.8 & \cellcolor{cyan!20}\textbf{73.3} & 73.0 & 72.5 & 72.2 \\
\bottomrule
\end{tabular}
\label{tab:ablation}
%}
\end{table*}%
%%%%%%%%%%% NEW TABLE %%%%%%%%%%%%%%

\subsection{Few-shot transfer}
\label{sec:fewshot}
The results of the few-shot transfer are shown in \Cref{tab:few_shot}. Conclusions analogous to \Cref{sec:imagenet-benchmark} can be drawn: AdaSim improves the downstream performance on most datasets and on average (last column). For a point of comparison, \Cref{tab:few_shot} contains a row ``Supervised'' taken from \cite{ericsson_how_2021} which is obtained with the weights from the supervised ResNet-50 in torchvision. Interestingly, DINO-2+AdaSim performs much better than the baseline DINO-2 on datasets where the supervised method also performs better \eg Cars \cite{krause_collecting_nodate} (+5.06) or Aircraft \cite{aircraft_dataset} (+2.54). This is because AdaSim bootstraps neighbors in the latent space which acts as a sort of self-labeling and therefore shares some properties with the supervised method.

\begin{figure*}[b]
\vspace{-0.1cm}
    \centering
    \begin{minipage}{.33\linewidth}
    \centering
    \subcaptionbox{Neighbor bootstrapping ratio \label{fig:1a} }
    {\includegraphics[width=\textwidth]{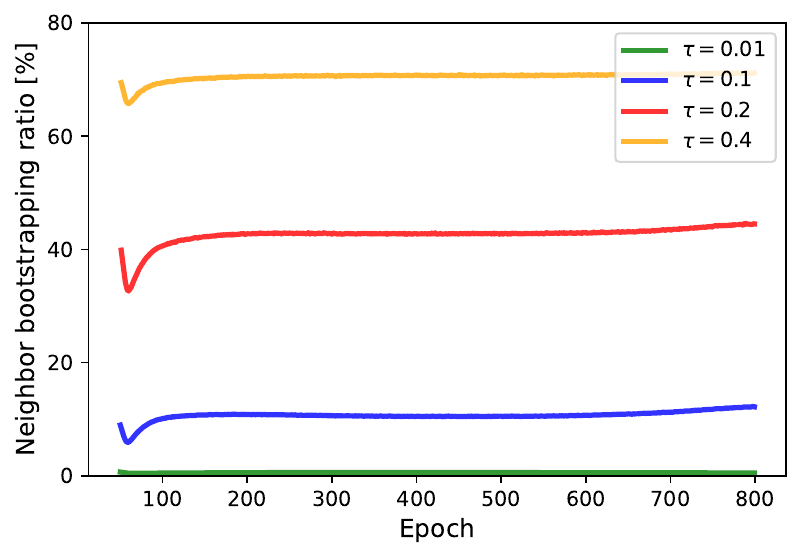}}
    \end{minipage}
    \begin{minipage}{.33\linewidth}
    \centering
    \subcaptionbox{NN top-1 training accuracy \label{fig:1b} }
    {\includegraphics[width=\textwidth]{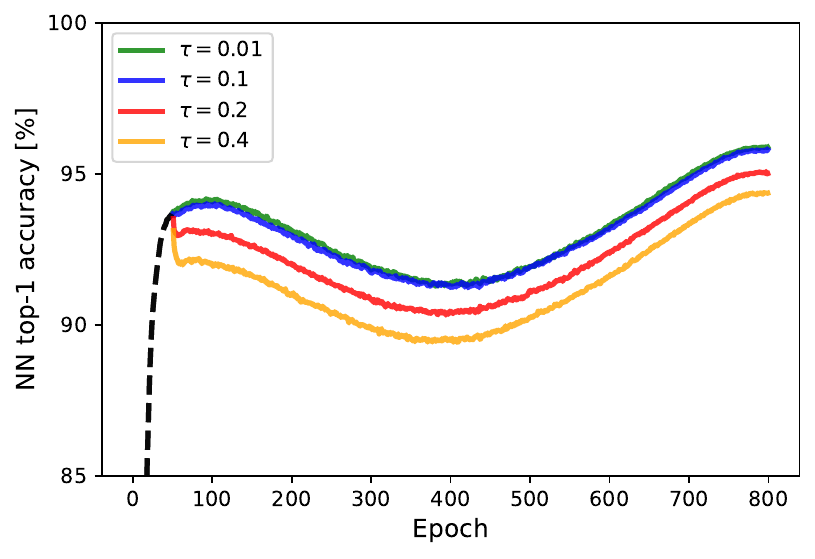}}
    \end{minipage}
    \begin{minipage}{.33\linewidth}
    \centering
    \subcaptionbox{2-NN top-1 training accuracy \label{fig:1c}}
    {\includegraphics[width=\textwidth]{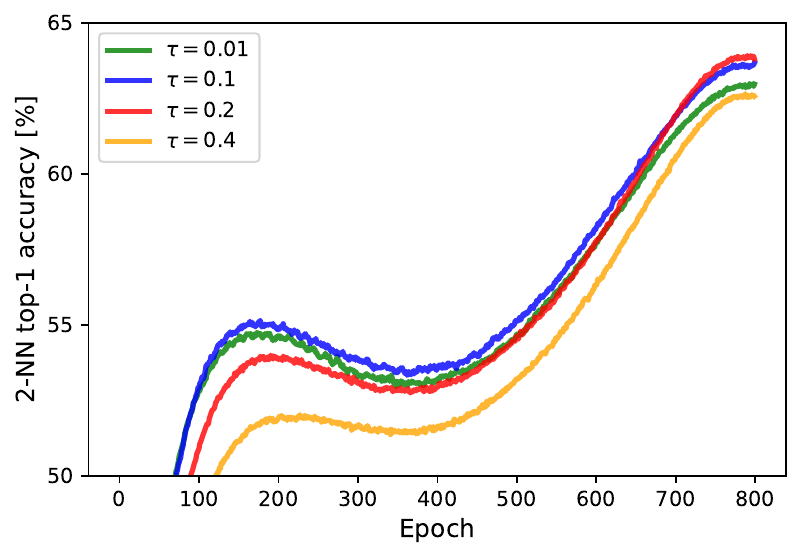}}
    \end{minipage}
    \caption{\textbf{Visualization of multiple training metrics for different temperature $\tau$ values.}}
    \label{fig:plots_under_hood}
\end{figure*}

\subsection{Ablations}
\label{sec:ablations}
An ablation study over AdaSim specific hyperparameters ($\tau$, $w$, $K$) can be found in \Cref{tab:ablation}. The best hyperparameters are highlighted in bold. These bold parameters are used for all runs, except for the parameter that is being varied. Importantly, \textbf{for a temperature $\tau=0$, AdaSim behaves like a standard self-distillation} method using positive pairs of the form $(t(\im), t'(\im))$. A performance improvement can be observed for increasing temperature values which shows the merits of AdaSim.

\subsection{Under the hood analysis}
\label{sec:underthehood}
Multiple training metrics are shown in \Cref{fig:plots_under_hood} with varying temperature values. Such plots are useful to build intuition on the internal mechanisms of AdaSim.

\noindent
{\bf Neighbor bootstrapping ratio} indicates the percentage of positive pairs $(t(\im_i),t'(\im_{j^\star}))$ where the augmentations are from different images. The higher the temperature, the higher the percentage is. In the limit when $\tau \to 0$, it can be observed that this percentage goes to 0, and AdaSim defaults to standard self-distillation. This is only possible thanks to the adaptive sampling of positive pairs in AdaSim (lines 13 to 18 in \Cref{alg:adasim}). Without the adaptive sampling, a low temperature would lead to a positive pair $(t(\im_i),t'(\im_{j^\star}))$ where $\im_{j^\star} = \argmax p^{\text{win}}(\im_j | \im_i)$ but there is no guarantee that $\im_i = \im_{j^\star}$. The adaptivity of the proposed method can be observed in \Cref{fig:plots_under_hood}\textcolor{red}{.a}. Indeed, at epoch 50, the window is filled and nearest neighbor bootstrapping is allowed to occur. As the quality of the latent space is low, so is that of the resulting gradients, which temporarily hurts the learned representations. Thanks to the adaptivity criterion, the bootstrapping ratio is automatically reduced to avoid collapse.
% Interestingly, we can observe the effect of switching from standard positive pairs to the adaptive setting at epoch 50 ($w=50$ in this case). When using positive pairs of the form $(t(\im_i),t'(\im_{j^\star}))$, they enforce gradients which are such that the $\argmax$ of $p^{\text{win}}(\im_j | \im_i)$ changes and therefore, less bootstrapping occurs due to the adaptivity.

\noindent
{\bf NN top-1 training accuracy} shows how often the query image $\im_i$ and its ``nearest neighbor'' $\im_{j^\star}$ are from the same class. Here we observe that a higher temperature leads to a lower accuracy which makes sense because the ``nearest neighbor'' can be the same image and, therefore, would trivially be in the same class. Note that before epoch 50, all temperature values use the same positive pairs as $w=50$ similarity metrics are being computed.

\noindent
{\bf 2-NN top-1 training accuracy} shows if the query image $\im_i$ and its second ``nearest neighbor'' $\argmax_{\im_j \neq \im_{j^\star}} p^{\text{win}}(\im_j | \im_i)$ are from the same class. This metric is a better indicator of the downstream generalizability of the learned features. It can be observed that higher temperature values (more neighbor bootstrapping) are initially worse but start to become advantageous as the training progresses. This is intuitive because bootstrapping neighbors is only useful when they are semantically related, which only happens as the network learns.

\noindent
{\bf Visualization of positive pairs} To get an understanding of the positive pairs which are formed by AdaSim, we visualize multiple query images $\im_i$ along with the sampling distribution $p^{\text{win}}(\im_j | \im_i)$ (overlayed in green) and its associated support $\mathcal{S}_{\text{win}}$ in \Cref{fig:query_support}. In this example, all ``nearest neighbors'' are the same as the query image, and all neighbors seem to share semantic content. In \Cref{sec:visualization}, we explicitly search for query images where the neighbors are from different classes. These results show evidence of wrongly labeled or duplicate images in ImageNet-1k \cite{imagenet}.

\section{Conclusion}
\label{sec:conclusion}
Self-distillation is becoming the go-to self-supervised learning paradigm due to its simplicity and state-of-the-art performance. However, non-explicit processing of negative pairs makes it less robust and more prone to collapse to trivial solutions than contrastive learning. Used in conjunction with bootstrapped positive pairs of neighbors, we empirically observe that self-distillation methods can perform worse than their vanilla baseline and in some cases even collapse. We propose an adaptive bootstrapping scheme that stabilizes the training and improves on the baselines. We also observe that long training schedules and larger backbones are particularly beneficial for AdaSim (better representations lead to better bootstrapping).

\noindent
{\bf Limitations} 
\label{par:limitations}
All results in the paper do not include multi-crop \cite{swav} for simplicity. In practice, not using multi-crop requires the use of more diverse random cropping (\eg with scale sampled in $[0.1, 1]$) but we have not changed any hyperparameters from DINO and stuck with $[0.25, 1]$.

\section*{Acknowledgement}
\label{sec:aknowledgement}
This project is funded by the European Research Council (ERC) under the European Union’s Horizon 2020 research and innovation program (Grant Agreement No. 101021347). This work is also partially funded by the Personalized Health and Related Technologies (PHRT), grant number 2021/344. We acknowledge EuroCC Belgium for awarding this project access to the LUMI supercomputer, owned by the EuroHPC Joint Undertaking, hosted by CSC (Finland) and the LUMI consortium.

{\small
\bibliographystyle{ieee_fullname}
\bibliography{references}
}

\clearpage
\newpage
\appendix
\setcounter{table}{0}
\renewcommand{\thetable}{A\arabic{table}}
\setcounter{figure}{0}
\renewcommand{\thefigure}{A\arabic{figure}}

\setcounter{equation}{0}
\renewcommand{\theequation}{A\arabic{equation}}

\section*{Appendix}
We provide additional details in \Cref{sec:abs}, \Cref{sec:implementation_details_appendix}, \Cref{sec:runtime} and \Cref{sec:scalability_app} as well as some additional quantitative results in \Cref{sec:additional_results_appendix} and qualitative visualizations in \Cref{sec:visualization}.

\section{Self-distillation asymmetry abstraction}
\label{sec:abs}
In \Cref{sec:selfdistillation-contrastive}, we state that self-distillation methods avoid collapse by using asymmetry and claim that this asymmetry can be abstracted out using two asymmetric encoders $f$ and $f'$. Given a distance metric $d$, a self-distillation objective is made solely out of positive terms of the form:

\begin{equation}
    \mathcal{L}_{distil} = d(f(\im), f'(\im^+)) \tag{\ref{eq:self-distil}}
\end{equation}
where $(\im, \im^+)$ is a positive pair. The total self-distillation objective is \Cref{eq:self-distil} summed over all positive pairs $(\im, \im^+)$. Below, we explicit the form of the encoders and the distance metric, both for SimSiam \cite{simsiam} and DINO \cite{dino}.

\subsection{SimSiam \cite{simsiam}}
Using notation from the original paper, SimSiam defines an encoder $f \colon \mathcal{X} \to \mathcal{Z}$ and a predictor $p\colon \mathcal{Z} \to \mathcal{Z}$. Using our notations, we encapsulate both the original encoder $f$ and the predictor $p$ in a single encoder which we also denote by $f$ and define our $f'$ as the $f$ from SimSiam\footnote{the notation on the right side of \Cref{eq:1} and \Cref{eq:2} refers to notation from SimSiam, and the left side refers to our notation}:
\begin{equation}
    f' \triangleq f
    \label{eq:1}
\end{equation}

\begin{equation}
    f \triangleq p \circ f
    \label{eq:2}
\end{equation}
The distance metric used is the negative cosine similarity. Given the above, the self-distillation loss in SimSiam can be written as
\begin{equation}
    d(f(\im), f'(\im^+)) := -\frac{f(\im)^\top f'(\im^+)}{\norm{f(\im)}_2 \norm{f'(\im^+)}_2}
\end{equation}
This loss is minimized w.r.t. the weights of $f$ (no gradients are back-propagated through $f'$).

\subsection{DINO \cite{dino}}
Using notation from the original paper, DINO uses a student backbone $g_{\theta_s} \colon \mathcal{X} \to \mathcal{P}$ where $\mathcal{P}$ is the space of discrete probability mass functions. An analogous teacher backbone $g_{\theta_t}$ is defined as a smoothed version of $g_{\theta_s}$. At the end of each epoch, the weights of the teacher backbone are updated with $\theta_t \leftarrow \lambda \theta_t + (1-\lambda) \theta_s$ where $\theta_s$ and $\theta_t$ refer to the weights of the student and teacher backbone, respectively.

Using our notations, we define both encoders as
\begin{equation}
    f \triangleq g_{\theta_s}
\end{equation}
and
\begin{equation}
    f' \triangleq g_{\theta_t}
\end{equation}
The distance metric $d$ is the cross entropy. Given the above, the self-distillation loss of DINO can be written as
% In DINO \cite{dino}, $f'$ is a momentum encoder $f'(\im) := f_{\text{EMA}}(\im)$ whose weights are updated at the end of each epoch as follows: $\weight' \leftarrow \lambda \weight' + (1-\lambda) \weight$. $\weight$ and $\weight'$ are the weights of $f$ and $f'$, respectively. Given a head $h \colon \mathcal{Z} \to \mathcal{P}$ mapping to the latent space discrete probability mass functions, and its exponential moving average equivalent $h'$, $d$ is defined as:

\begin{equation}
    d(f(\im), f'(\im^+)) := H(f(\im), f'(\im^+))
\end{equation}
where $H$ is the cross entropy
\begin{equation}
    H(p, q) = -\sum_{i \in \mathcal{I}} p(i) \log q(i)
\end{equation}
and $\mathcal{I}$ is the support of the distributions $p$ and $q$, \ie $\mathcal{I}=[I]=\{1, 2, \cdots, I\}$. $I$ refers to the dimensionality of the output distributions. This loss is minimized w.r.t. to the weights of $f$ (no gradients are back-propagated through $f'$).

\section{Implementation details}
\label{sec:implementation_details_appendix}
\subsection{SimSiam \cite{simsiam}}
We use the code from their official GitHub  (\href{https://github.com/facebookresearch/simsiam}{link}). All 3 runs (baseline, baseline + NN, baseline + Adasim) use the same hyperparameters:

\begin{itemize}
\setlength\itemsep{-0.3em}
    \item \texttt{arch}: resnet50
    \item \texttt{epochs}: 100
    \item \texttt{batch\_size}: 512
    \item \texttt{lr}: 0.05
    \item \texttt{momentum}: 0.9
    \item \texttt{weight\_decay}: 0.0001
    \item \texttt{dim}: 2048
    \item \texttt{pred\_dim}: 512
    \item \texttt{fix\_pred\_lr}: True
\end{itemize}

\subsection{DINO \cite{dino}}
We use the code from their official GitHub  (\href{https://github.com/facebookresearch/dino}{link}). For the 3 ResNet-50 runs (baseline, baseline + NN, baseline + Adasim) the same hyperparameters specified on the official GitHub are used, except for \texttt{local\_crops\_number} which we set to 0:

\begin{itemize}
\setlength\itemsep{-0.3em}
    \item \texttt{arch}: resnet50
    \item \texttt{batch\_size\_total}: 1024
    \item \texttt{clip\_grad}: 0.0
    \item \texttt{drop\_path\_rate}: 0.1
    \item \texttt{epochs}: 100
    \item \texttt{freeze\_last\_layer}: 1
    \item \texttt{global\_crops\_scale}: [0.14, 1.0]
    \item \texttt{local\_crops\_number}: 0
    \item \texttt{lr}: 0.03
    \item \texttt{lr\_linear}: 0.03
    \item \texttt{min\_lr}: 1e-05
    \item \texttt{momentum\_teacher}: 0.996
    \item \texttt{norm\_last\_layer}: False
    \item \texttt{optimizer}: sgd
    \item \texttt{out\_dim}: 65536
    \item \texttt{seed}: 0
    \item \texttt{teacher\_temp}: 0.07
    \item \texttt{use\_bn\_in\_head}: False
    \item \texttt{use\_fp16}: False
    \item \texttt{warmup\_epochs}: 10
    \item \texttt{warmup\_teacher\_temp}: 0.04
    \item \texttt{warmup\_teacher\_temp\_epochs}: 30
    \item \texttt{weight\_decay}: 0.0001
    \item \texttt{weight\_decay\_end}: 0.0001
\end{itemize}

\begin{figure}[b]
    \centering
    \includegraphics[width=1\columnwidth]{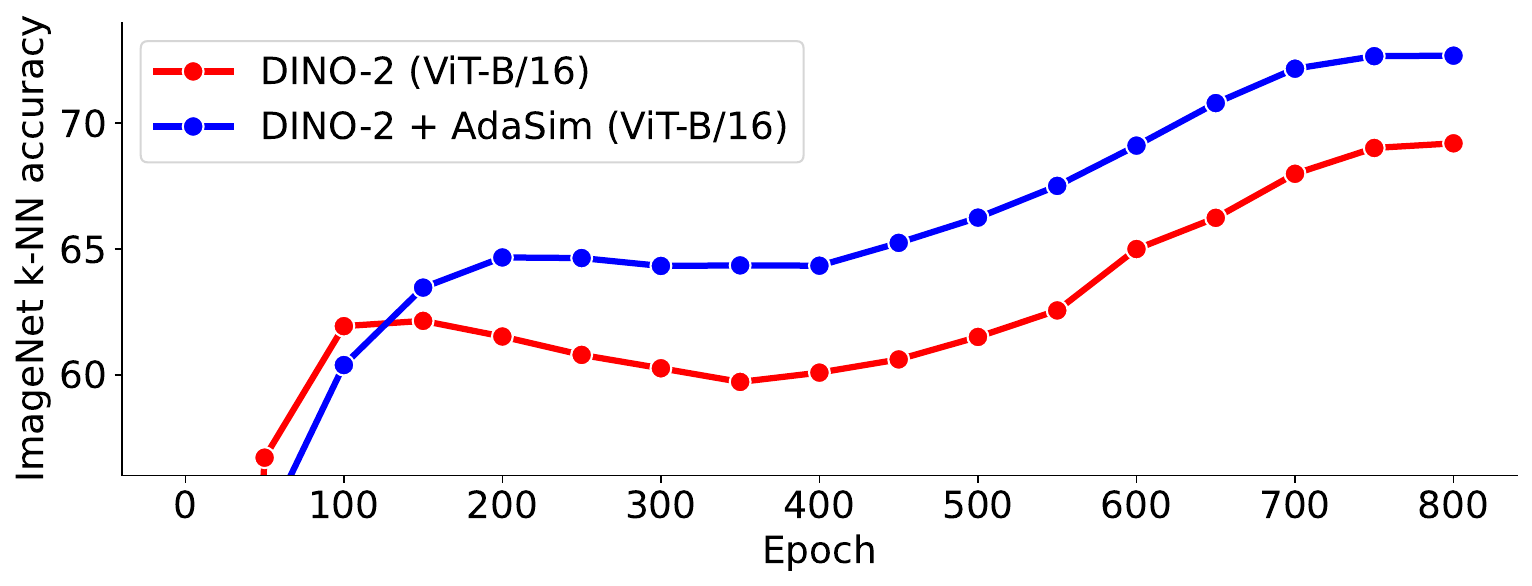}
    \caption{AdaSim benefits from \textbf{longer training schedules}.}
    \label{fig:longer_training}
\end{figure}

For all runs using the ViT-S/16 backbone, the same hyperparameters specified on the official GitHub are used:

\begin{itemize}
\setlength\itemsep{-0.3em}
    \item \texttt{arch}: vit\_small
    \item \texttt{patch\_size}: 16
    \item \texttt{batch\_size\_total}: 1024
    \item \texttt{clip\_grad}: 0.0
    \item \texttt{drop\_path\_rate}: 0.1
    \item \texttt{epochs}: 800
    \item \texttt{freeze\_last\_layer}: 1
    \item \texttt{global\_crops\_scale}: [0.4, 1.0]
    \item \texttt{local\_crops\_number}: 0
    \item \texttt{lr}: 0.0005
    \item \texttt{min\_lr}: 1e-05
    \item \texttt{momentum\_teacher}: 0.996
    \item \texttt{norm\_last\_layer}: False
    \item \texttt{optimizer}: adamw
    \item \texttt{out\_dim}: 65536
    \item \texttt{seed}: 0
    \item \texttt{teacher\_temp}: 0.07
    \item \texttt{use\_bn\_in\_head}: False
    \item \texttt{use\_fp16}: True
    \item \texttt{warmup\_epochs}: 10
    \item \texttt{warmup\_teacher\_temp}: 0.04
    \item \texttt{warmup\_teacher\_temp\_epochs}: 30
    \item \texttt{weight\_decay}: 0.04
    \item \texttt{weight\_decay\_end}: 0.4
\end{itemize}

\section{Runtime analysis}
\label{sec:runtime}
We compare the runtime for baseline DINO-2 \cite{dino} and DINO-2 + AdaSim. As can be seen in \Cref{tab:timings}, AdaSim has a negligible impact on throughput during the pretraining.

\section{Scalability}
\label{sec:scalability_app}
The memory required to store the cache $\Z$ is linear in the size of the dataset. However, its footprint is always much lower than the dataset itself since the cache only stores a representation instead of a full image. When training on very large datasets, workers do not store the whole dataset but only a shard $\mathcal{D}^{(i)}$ with $\mathcal{D} = \{\mathcal{D}^{(1)}, \mathcal{D}^{(2)}, \cdots \mathcal{D}^{(n)}\}$. In such case, the cache $\Z$ can also be split into shards $\Z^{(i)}$ with $\Z = \{\Z^{(1)}, \Z^{(2)}, \cdots \Z^{(n)}\}$ making the algorithm scalable to datasets of arbitrary size.

\begin{table}[t]
\caption{\textbf{Runtime analysis of AdaSim per iteration of pretraining.} Run on 4x AMD MI250X GPUs.}
\resizebox{\columnwidth}{!}{
\begin{tabular}{lccc}
\toprule
method & backbone & batchsize per GPU & time per iter [s] \\
\midrule
DINO-2 \cite{dino} & ViT-S/16 & 128 & 0.256 \\
\rowcolor{cyan!20} DINO-2 + AdaSim & ViT-S/16 & 128 & 0.270\\

\midrule
DINO-2 \cite{dino} & ViT-S/16 & 256 & 0.480 \\
\rowcolor{cyan!20} DINO-2 + AdaSim & ViT-S/16 & 256 & 0.488\\

\bottomrule
\end{tabular}
}
\label{tab:timings}
\end{table}

\section{Additional results}
\label{sec:additional_results_appendix}
\begin{table}[t]
\centering
\caption{
\textbf{Transfer to linear segmentation.} A linear layer is trained on top of the frozen spatial features. We report mIoU scores on PVOC12, COCO-Thing, and COCO-Stuff. ``\collapse'' denotes collapse.
}
\footnotesize
\resizebox{1\columnwidth}{!}{
\begin{tabular}{l c c c c c c c c c c c c}
\toprule
&&&&& \multicolumn{4}{c}{mIoU scores} \\
\cmidrule{6-9}
Method & Model & Dataset & Epochs && PVOC12 & COCO-Thing & COCO-Stuff & Avg.\\
% \midrule
\midrule
DINO-2  & ViT-S/16 & ImageNet-1k & 800 && 69.0 & 65.0 & \textbf{52.4} & 62.1\\
DINO-2 + NN  & ViT-S/16 & ImageNet-1k & 800 && \collapse & \collapse & \collapse & \collapse\\
\rowcolor{cyan!20} DINO-2 + AdaSim & ViT-S/16 & ImageNet-1k  & 800  && \textbf{69.6} & \textbf{65.7} & 52.0 & \textbf{62.4} \textcolor{darkgreen}{(+ 0.3)}\\
% \midrule
% DINO-2  & ViT-B/16 & ImageNet & 800 &&  &  &  & \\
% DINO-2 + NN  & ViT-B/16 & ImageNet & 800 && \collapse & \collapse & \collapse & \collapse\\
% \rowcolor{cyan!20} DINO-2 + AdaSim & ViT-B/16 & ImageNet  & 800  &&  &  &  & \\
\midrule
% DINO-2  & ViT-S/16  & COCO & 400 && 47.6 & 48.2 & 46.5 \\
DINO-2  & ViT-S/16  & COCO & 800 && 49.9 & 48.1 & 46.3 & 48.1\\
DINO-2 + NN  & ViT-S/16 & COCO & 800 && \collapse & \collapse & \collapse & \collapse\\
% \rowcolor{cyan!20} AdaSim & ViT-S/16  & COCO  & 400 && \textbf{48.5} & \textbf{49.0} & \textbf{46.9}  \\
\rowcolor{cyan!20} DINO-2 + AdaSim & ViT-S/16  & COCO  & 800 && \textbf{52.9} & \textbf{52.2} & \textbf{48.1} & \textbf{51.1} \textcolor{darkgreen}{(+ 3.0)}\\

\bottomrule
\end{tabular}
\label{table:linear_segmentation}
}
\end{table}%

% ==> 230529_ablation_vote50_k3_t02_vitb16_400_coco/LC_VOCSegmentation_4/mIoU_results.txt <==
% mIoU: 48.47107076512096

% ==> 230529_ablation_vote50_k3_t02_vitb16_400_coco/LC_coco_thing_4/mIoU_results.txt <==
% mIoU: 48.97735107405059

% ==> 230529_ablation_vote50_k3_t02_vitb16_400_coco/LC_coco_stuff_4/mIoU_results.txt <==
% mIoU: 46.941400047591245

We investigate the ability of AdaSim to cope with scene-centric datasets and to produce spatial features aligned with dense downstream tasks. To that end, we train a ViT-S/16 for 800 epochs on COCO~\cite{lin2014microsoft}. The resulting features are then evaluated via a linear segmentation task on three datasets, namely PVOC12~\cite{pascal-voc-2012}, COCO-Thing, and COCO-Stuff~\cite{lin2014microsoft}. We rely on the evaluation pipeline of \cite{stegmuller2023croc} and refer to their work for the implementation details.

%\Tim{Additionally, we evaluate the performance of AdaSim compared to the baseline over the training epochs in \Cref{fig:longer_training}. The reported metric is the $k$-NN accuracy on ImageNet \cite{imagenet}. We observe that AdaSim is initially slower than the baseline but quickly outperforms it. This is because the learned representations must be of good quality for any bootstrapping to be beneficial. Bootstrapping pairs of nearest neighbors that aren’t semantically related hurts the performance and is part of the motivation behind AdaSim. Both larger backbones and longer training schedules contribute to learning better representations.}
Additionally, we examine how the performance of AdaSim evolves throughout training. For that purpose, we report the $k$-NN accuracy on ImageNet \cite{imagenet} at different epochs. We observe that AdaSim initially performs worse than the baseline (DINO-2 + ViT-B/16), but outperforms it from $\approx 150$ epochs on. Indeed, the learned representations must be sufficiently good for bootstrapping to be beneficial. Bootstrapping pairs of nearest neighbors that aren’t semantically related hurts the performance and is part of the motivation behind AdaSim. Along the same line, we observe that both larger backbones and longer training schedules contribute to learning better representations.

\section{Investigating intriguing neighbors}
\label{sec:visualization}
In \Cref{fig:query_support}, we show a visualization of random query images $\im_i$ and their corresponding support $\mathcal{S}_{\text{win}}$ ranked by decreasing $p^{\text{win}}(\im_j | \im_i)$. All 1-NN are the same as the query image ($\im_i = \argmax_{\im_j}p^{\text{win}}(\im_j | \im_i)$), and all other neighbors seem to share semantic content with the query image.

\subsection{Nearest neighbor is different from the query}
\begin{figure*}[t]
    \centering
    \includegraphics[width=\textwidth]{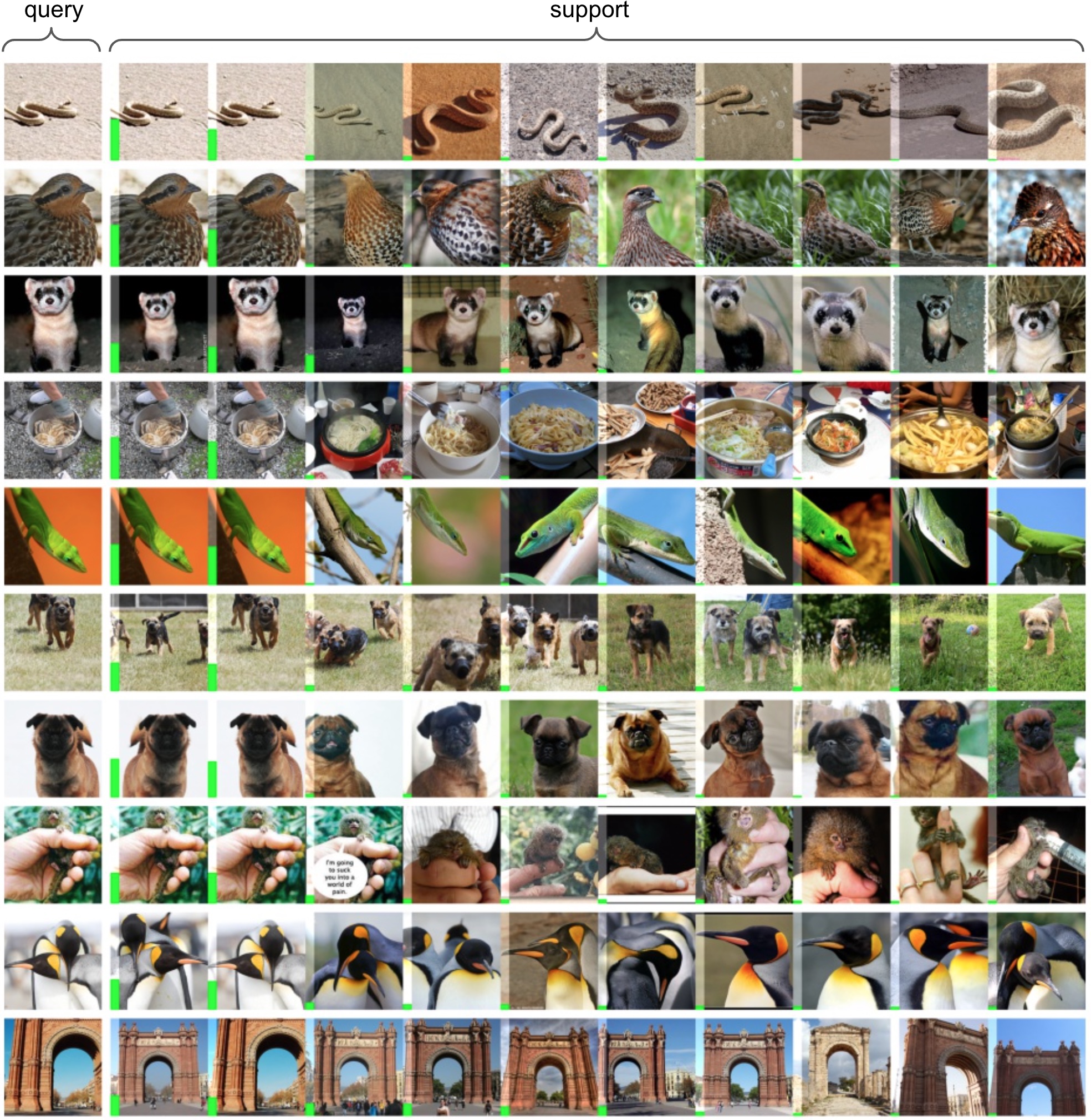}
    \caption{\textbf{Query image $\im_i$ (left column) and corresponding support $\mathcal{S}_{\text{win}}$ ranked by decreasing $p^{\text{win}}(\im_j | \im_i)$ (illustrated as a \textcolor{green}{green} bar on the bottom left of each image).} The query images are chosen such that the most similar image from the support is not the same as the query ($\im_i \neq \argmax_{\im_j}p^{\text{win}}(\im_j | \im_i)$). Metadata about the images is shown in \Cref{tab:firstisnotid}.}
    \label{fig:firstisnotid}
\end{figure*}

Here, we explicitly search for cases where the nearest neighbor is not the same image as the query to observe border cases of AdaSim. Mathematically, this is the case when $\im_i \neq \argmax_{\im_j}p^{\text{win}}(\im_j | \im_i)$. Such queries are shown in \Cref{fig:firstisnotid} with the corresponding metadata in \Cref{tab:firstisnotid}. It can be observed that even when the nearest neighbor does not originate from the same image, all nearest neighbors visually share semantic content. 

\subsection{Nearest neighbor is from a different class as the query}
\begin{figure*}[t]
    \centering
    \includegraphics[width=\textwidth]{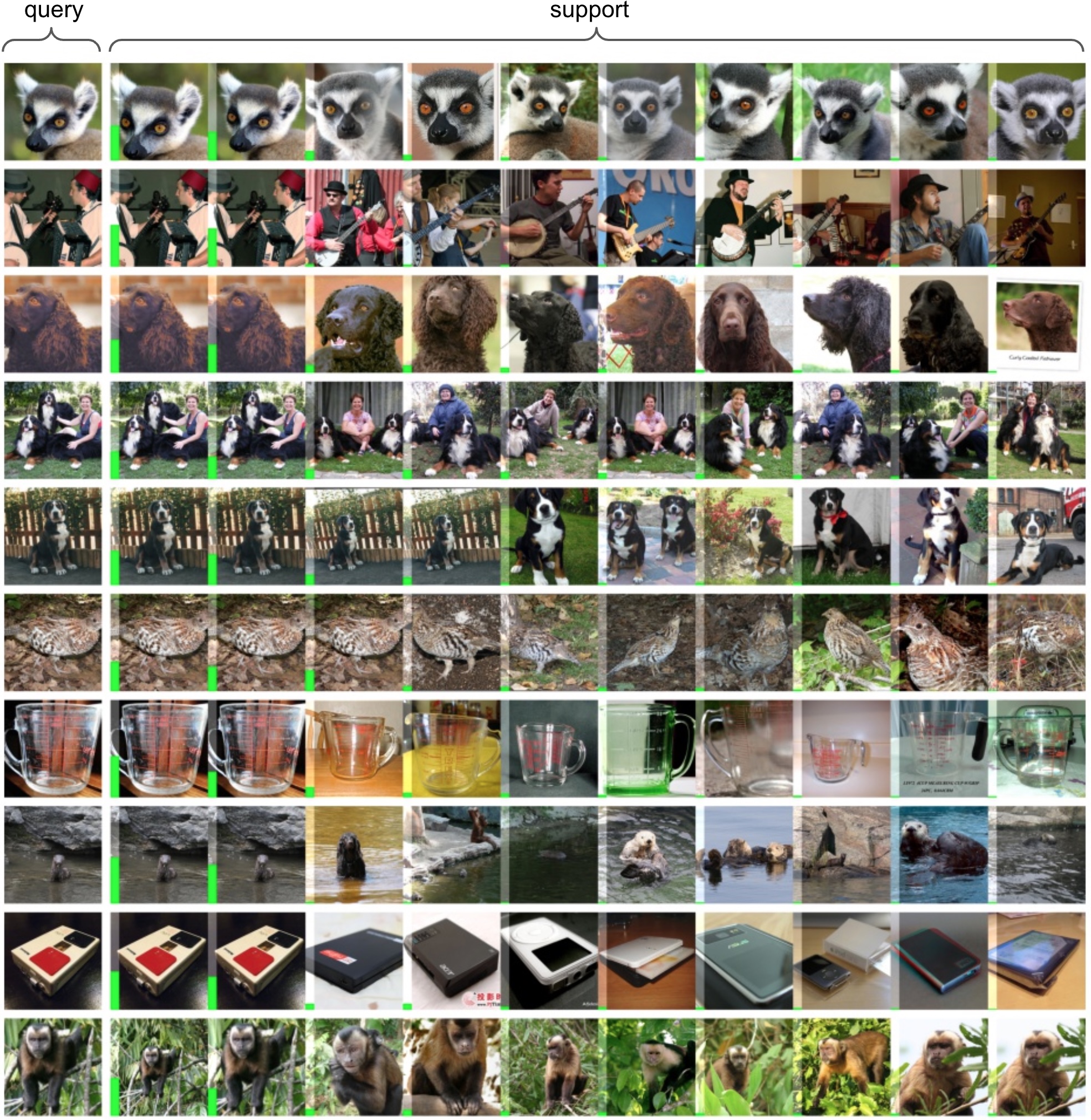}
    \caption{\textbf{Query image $\im_i$ (left column) and corresponding support $\mathcal{S}_{\text{win}}$ ranked by decreasing $p^{\text{win}}(\im_j | \im_i)$ (illustrated as a \textcolor{green}{green} bar on the bottom left of each image).} The query images are chosen such that the most similar image from the support is not from the same class as the query ($\texttt{class}(\im_i) \neq \texttt{class}(\argmax_{\im_j}p^{\text{win}}(\im_j | \im_i))$). Metadata about the images is shown in \Cref{tab:firstisnotclass}.}
    \label{fig:firstisnotclass}
\end{figure*}

A stronger special case happens when the nearest neighbor is not even from the same class as the query \ie $\texttt{class}(\im_i) \neq \texttt{class}(\argmax_{\im_j}p^{\text{win}}(\im_j | \im_i))$. This is shown in \Cref{fig:firstisnotclass} and \Cref{tab:firstisnotclass}. Interestingly, even if the first nearest neighbor is not from the same class, it still looks very similar. This shows evidence of mislabelling. Consider the example of the first row of \Cref{fig:firstisnotclass}. The query is from class 384 (indri, indris, Indri indri, Indri brevicaudatus) yet the first neighbor is from class 383 (Madagascar cat, ring-tailed lemur, Lemur catta). However, they are very similar and one seems to be a zoomed-in version of the other. Similar conclusions can be drawn for all other query images in \Cref{fig:firstisnotclass}.\\

For readability purposes, \Cref{tab:firstisnotid} and \Cref{tab:firstisnotclass} only include the class ids and not the full class names. A mapping from class id to class name can be found below for the subset of classes appearing in \Cref{tab:firstisnotid} and \Cref{tab:firstisnotclass}.

\begin{itemize}
\setlength\itemsep{-0.5em}
\item 60: night snake, Hypsiglena torquata
\item 66: horned viper, cerastes, sand viper, horned asp, Cerastes cornutus
\item 68: sidewinder, horned rattlesnake, Crotalus cerastes
\item 80: black grouse
\item 82: ruffed grouse, partridge, Bonasa umbellus
\item 83: prairie chicken, prairie grouse, prairie fowl
\item 138: bustard
\item 166: Walker hound, Walker foxhound
\item 167: English foxhound
\item 206: curly-coated retriever
\item 219: cocker spaniel, English cocker spaniel, cocker
\item 220: Sussex spaniel
\item 221: Irish water spaniel
\item 238: Greater Swiss Mountain dog
\item 239: Bernese mountain dog
\item 240: Appenzeller
\item 241: EntleBucher
\item 244: Tibetan mastiff
\item 337: beaver
\item 341: hog, pig, grunter, squealer, Sus scrofa
\item 342: wild boar, boar, Sus scrofa
\item 343: warthog
\item 356: weasel
\item 357: mink
\item 358: polecat, fitch, foulmart, foumart, Mustela putorius
\item 359: black-footed ferret, ferret, Mustela nigripes
\item 360: otter
\item 365: orangutan, orang, orangutang, Pongo pygmaeus
\item 370: guenon, guenon monkey
\item 376: proboscis monkey, Nasalis larvatus
\item 378: capuchin, ringtail, Cebus capucinus
\item 380: titi, titi monkey
\item 383: Madagascar cat, ring-tailed lemur, Lemur catta
\item 384: indri, indris, Indri indri, Indri brevicaudatus
\item 386: African elephant, Loxodonta africana
\item 401: accordion, piano accordion, squeeze box
\item 420: banjo
\item 482: cassette player
\item 487: cellular telephone, cellular phone, cellphone, cell, mobile phone
\item 546: electric guitar
\item 574: golf ball
\item 592: hard disc, hard disk, fixed disk
\item 605: iPod
\item 647: measuring cup
\item 745: projector
\item 819: stage
\item 848: tape player
\item 852: tennis ball
\item 863: totem pole
\item 890: volleyball
\item 968: cup

\end{itemize}

\begin{table*}[t]
\caption{\textbf{Metadata corresponding to \Cref{fig:firstisnotid}.} Each block corresponds to a row of \Cref{fig:firstisnotid}. Within a block, the first row denotes the image id, the second row denotes the class id and the last row denotes the sampling distribution $p^{\text{win}}(\im_j | \im_i)$.}
\resizebox{\textwidth}{!}{
\begin{tabular}{ccccccccccc}
\toprule
$\im_i$ (query)   & 1-NN    & 2-NN    & 3-NN    & 4-NN    & 5-NN    & 6-NN    & 7-NN    & 8-NN    & 9-NN    & 10-NN   \\
\midrule
313037  & 313303  & 313037  & 312150  & 312728  & 312960  & 312198  & 313277  & 312158  & 312230  & 312903  \\
244     & 244     & 244     & 244     & 244     & 244     & 244     & 244     & 244     & 244     & 244     \\
       & 0.41    & 0.36    & 0.04    & 0.04    & 0.04    & 0.03    & 0.02    & 0.02    & 0.02    & 0.02    \\
\midrule
1105805 & 1105953 & 1105805 & 1105673 & 1105826 & 1106178 & 1106303 & 1105027 & 1105739 & 1105747 & 1105980 \\
863     & 863     & 863     & 863     & 863     & 863     & 863     & 863     & 863     & 863     & 863     \\
       & 0.2     & 0.16    & 0.16    & 0.14    & 0.13    & 0.07    & 0.06    & 0.04    & 0.02    & 0.02    \\
\midrule
1140581 & 1139752 & 1140581 & 1139966 & 1139623 & 1140694 & 1140389 & 1139643 & 1140611 & 1140717 & 1139731 \\
890     & 890     & 890     & 890     & 890     & 890     & 890     & 890     & 890     & 890     & 890     \\
       & 0.21    & 0.18    & 0.14    & 0.11    & 0.1     & 0.09    & 0.06    & 0.06    & 0.03    & 0.02    \\
\midrule
213504  & 214625  & 213504  & 214919  & 213989  & 214090  & 213530  & 214654  & 214352  & 214537  & 214521  \\
166     & 167     & 166     & 167     & 166     & 166     & 166     & 167     & 167     & 167     & 167     \\
& 0.2     & 0.15    & 0.13    & 0.13    & 0.11    & 0.11    & 0.09    & 0.06    & 0.02    & 0.01    \\
\midrule
482666  & 483351  & 482666  & 482655  & 482983  & 482561  & 482824  & 483621  & 482626  & 482600  & 482467  \\
376     & 376     & 376     & 376     & 376     & 376     & 376     & 376     & 376     & 376     & 376     \\
& 0.3     & 0.29    & 0.1     & 0.06    & 0.06    & 0.05    & 0.03    & 0.03    & 0.03    & 0.03    \\
\midrule
439007  & 437919  & 439007  & 439041  & 440492  & 440579  & 496403  & 496313  & 179490  & 440440  & 439580  \\
342     & 341     & 342     & 342     & 343     & 343     & 386     & 386     & 138     & 343     & 343     \\
& 0.27    & 0.24    & 0.13    & 0.07    & 0.07    & 0.05    & 0.05    & 0.04    & 0.04    & 0.04    \\
\midrule
304802  & 307986  & 304802  & 305026  & 308855  & 308662  & 305875  & 305635  & 307400  & 305792  & 307371  \\
238     & 240     & 238     & 238     & 241     & 241     & 239     & 239     & 240     & 239     & 240     \\
& 0.32    & 0.2     & 0.11    & 0.09    & 0.09    & 0.08    & 0.04    & 0.03    & 0.03    & 0.02    \\
\midrule
86529   & 88177   & 86529   & 88271   & 88804   & 78641   & 88860   & 88405   & 86130   & 89098   & 86108   \\
66      & 68      & 66      & 68      & 68      & 60      & 68      & 68      & 66      & 68      & 66      \\
& 0.42    & 0.41    & 0.03    & 0.03    & 0.02    & 0.02    & 0.02    & 0.02    & 0.01    & 0.01    \\
\midrule
738207  & 1091457 & 738207  & 738128  & 738179  & 738186  & 737971  & 737433  & 737058  & 737339  & 738233  \\
574     & 852     & 574     & 574     & 574     & 574     & 574     & 574     & 574     & 574     & 574     \\
& 0.23    & 0.21    & 0.14    & 0.12    & 0.08    & 0.05    & 0.05    & 0.05    & 0.04    & 0.04    \\
\midrule
460081  & 461411  & 460081  & 461155  & 460028  & 459517  & 461393  & 459646  & 460305  & 459219  & 460791  \\
358     & 359     & 358     & 359     & 358     & 358     & 359     & 358     & 359     & 358     & 359     \\
& 0.37    & 0.35    & 0.08    & 0.05    & 0.04    & 0.03    & 0.02    & 0.02    & 0.02    & 0.02   \\
\bottomrule
\end{tabular}
}
\label{tab:firstisnotid}
\end{table*}

\begin{table*}[]
\caption{\textbf{Metadata corresponding to \Cref{fig:firstisnotclass}.} Each block corresponds to a row of \Cref{fig:firstisnotclass}. Within a block, the first row denotes the image id, the second row denotes the class id and the last row denotes the sampling distribution $p^{\text{win}}(\im_j | \im_i)$.}
\resizebox{\textwidth}{!}{
\begin{tabular}{ccccccccccc}
\toprule
$\im_i$ (query)   & 1-NN    & 2-NN    & 3-NN    & 4-NN    & 5-NN    & 6-NN    & 7-NN    & 8-NN    & 9-NN    & 10-NN   \\
\midrule
493896 & 492056  & 493896 & 492267 & 493504 & 492574 & 492098  & 491860 & 493914 & 493160 & 492988 \\
384    & 383     & 384    & 383    & 384    & 383    & 383     & 383    & 384    & 384    & 384    \\
& 0.36    & 0.3    & 0.1    & 0.06   & 0.04   & 0.04    & 0.03   & 0.03   & 0.03   & 0.02   \\
\midrule
514556 & 540102  & 514556 & 539549 & 539488 & 539085 & 1049500 & 540080 & 539722 & 539428 & 701031 \\
401    & 420     & 401    & 420    & 420    & 420    & 819     & 420    & 420    & 420    & 546    \\
& 0.43    & 0.4    & 0.03   & 0.03   & 0.02   & 0.02    & 0.02   & 0.02   & 0.02   & 0.02   \\
\midrule
281630 & 282778  & 281630 & 263441 & 282465 & 263810 & 263486  & 281511 & 282504 & 280302 & 263945 \\
220    & 221     & 220    & 206    & 221    & 206    & 206     & 220    & 221    & 219    & 206    \\
& 0.34    & 0.29   & 0.08   & 0.08   & 0.06   & 0.04    & 0.03   & 0.03   & 0.03   & 0.02   \\
\midrule
307055 & 309205  & 307055 & 308671 & 307104 & 308841 & 307955  & 308794 & 309282 & 309246 & 309227 \\
240    & 241     & 240    & 241    & 240    & 241    & 240     & 241    & 241    & 241    & 241    \\
& 0.29    & 0.23   & 0.1    & 0.1    & 0.07   & 0.07    & 0.05   & 0.04   & 0.03   & 0.02   \\
\midrule
304813 & 308718  & 304813 & 304687 & 307947 & 308807 & 308163  & 308745 & 307713 & 308216 & 307930 \\
238    & 241     & 238    & 238    & 240    & 241    & 241     & 241    & 240    & 241    & 240    \\
& 0.35    & 0.32   & 0.11   & 0.08   & 0.03   & 0.03    & 0.02   & 0.02   & 0.02   & 0.02   \\
\midrule
107449 & 107748  & 103927 & 107449 & 106473 & 107398 & 106690  & 106705 & 106982 & 106536 & 106610 \\
82     & 83      & 80     & 82     & 82     & 82     & 82      & 82     & 82     & 82     & 82     \\
& 0.31    & 0.26   & 0.22   & 0.06   & 0.05   & 0.04    & 0.03   & 0.02   & 0.01   & 0.01   \\
\midrule
830394 & 1239824 & 830394 & 831098 & 830660 & 831378 & 831425  & 830417 & 831372 & 831192 & 830232 \\
647    & 968     & 647    & 647    & 647    & 647    & 647     & 647    & 647    & 647    & 647    \\
& 0.43    & 0.27   & 0.11   & 0.05   & 0.04   & 0.03    & 0.02   & 0.02   & 0.02   & 0.02   \\
\midrule
458411 & 457058  & 458411 & 283078 & 469050 & 432009 & 462780  & 461777 & 457794 & 462065 & 461770 \\
357    & 356     & 357    & 221    & 365    & 337    & 360     & 360    & 357    & 360    & 360    \\
& 0.47    & 0.4    & 0.02   & 0.02   & 0.02   & 0.02    & 0.01   & 0.01   & 0.01   & 0.01   \\
\midrule
619775 & 1086664 & 619775 & 760794 & 954774 & 777417 & 1085978 & 626568 & 776959 & 760171 & 759785 \\
482    & 848     & 482    & 592    & 745    & 605    & 848     & 487    & 605    & 592    & 592    \\
& 0.4     & 0.34   & 0.07   & 0.07   & 0.03   & 0.03    & 0.02   & 0.02   & 0.01   & 0.01   \\
\midrule
488570 & 485197  & 488570 & 486127 & 488201 & 485483 & 485652  & 485828 & 484997 & 485594 & 475405 \\
380    & 378     & 380    & 378    & 380    & 378    & 378     & 378    & 378    & 378    & 370    \\
& 0.39    & 0.28   & 0.07   & 0.06   & 0.05   & 0.04    & 0.03   & 0.03   & 0.02   & 0.02  \\
\bottomrule
\end{tabular}
}
\label{tab:firstisnotclass}
\end{table*}

\end{document}